%% file: 00main.tex
\documentclass[acmlarge,screen,review=false]{acmart}

\AtBeginDocument{%
  \providecommand\BibTeX{{%
    \normalfont B\kern-0.5em{\scshape i\kern-0.25em b}\kern-0.8em\TeX}}}

\usepackage{multirow}
\usepackage{subfigure}
\usepackage{wrapfig}
\usepackage{xspace}
\usepackage{algorithm}
\usepackage{algpseudocode}
\usepackage{lipsum}
\usepackage{algorithm,algpseudocode,float}
\usepackage{subcaption}
\usepackage{graphicx}
\usepackage{makecell}
\usepackage[skip=8pt]{caption}

\begin{document}
\newcommand{\name}{PrivateGaze\xspace}
\renewcommand{\algorithmicrequire}{\textbf{Input:}}
\renewcommand{\algorithmicensure}{\textbf{Output:}}
\title{{\name}: Preserving User Privacy in Black-box Mobile Gaze Tracking Services}

% \author{Anonymous Author(s)}

\author{Lingyu Du}
\affiliation{
 \department{%Embedded Systems Group, 
 Department of Software Technology}
 \institution{Delft University of Technology}
 \city{Delft}
 \country{The Netherlands}
 }
\email{Lingyu.Du@tudelft.nl}

% Pennsylvania State University, State College, Pennsylvania, United States, jinyuan@psu.edu

\author{Jinyuan Jia}
\affiliation{%
\department{College of Information Sciences and Technology}
 \institution{The Pennsylvania State University}
 \city{State College}
 \country{The United States}}
\email{jinyuan@psu.edu}

\author{Xucong Zhang}
\affiliation{
\department{%Computer Vision Lab, 
Department of Intelligent Systems}
 \institution{Delft University of Technology}
 \city{Delft}
 \country{The Netherlands}}
 \email{xucong.zhang@tudelft.nl}

\author{Guohao Lan}
\affiliation{%
\department{%Embedded Systems Group, 
Department of Software Technology}
 \institution{Delft University of Technology}
 \city{Delft}
 \country{The Netherlands}}
 \email{g.lan@tudelft.nl}

\renewcommand{\shortauthors}{L. Du et al.}

\begin{abstract}
\input{01abstract}
\end{abstract}

\begin{CCSXML}
<ccs2012>
   <concept>
    <concept_id>10002978.10003029.10011150</concept_id>
       <concept_desc>Security and privacy~Privacy protections</concept_desc>
       <concept_significance>500</concept_significance>
       </concept>
   <concept>
       <concept_id>10003120.10003138</concept_id>
       <concept_desc>Human-centered computing~Ubiquitous and mobile computing</concept_desc>
       <concept_significance>500</concept_significance>
       </concept>
 </ccs2012>
\end{CCSXML}

\ccsdesc[500]{Security and privacy~Privacy protections}
\ccsdesc[500]{Human-centered computing~Ubiquitous and mobile computing}

\keywords{Mobile gaze estimation, black-box gaze tracking service, privacy preserving.}



\maketitle
\input{02introduction}
\input{03relatedwork}

\input{05method}
\input{06evaluation}

\input{07conclusion}

\begin{acks}
We would like to express our gratitude to the anonymous reviewers and the associate editors for their insightful comments and guidance. We also appreciate Koen G. Langendoen for his valuable comments and suggestions during our discussions. This work was supported in part by the Meta Research Award and by SURF Research Cloud grants EINF-2391, EINF-8964, and EINF-9272. The contents of this paper do not necessarily reflect the positions or policies of the funding agencies.
\end{acks}

\bibliographystyle{ACM-Reference-Format}
\bibliography{09reference}

\appendix

\end{document}

%% file: 01abstract.tex
Eye gaze contains rich information about human attention and cognitive processes. This capability makes the underlying technology, known as gaze tracking, a critical enabler for many ubiquitous applications and has triggered the development of easy-to-use gaze estimation services. Indeed, by utilizing the ubiquitous cameras on tablets and smartphones, users can readily access many gaze estimation services. In using these services, users must provide their full-face images to the gaze estimator, which is often a black box. This poses significant privacy threats to the users, especially when a malicious service provider gathers a large collection of face images to {classify} sensitive user attributes. In this work, we present \name, the first approach that can effectively preserve users' privacy in black-box gaze tracking services without compromising gaze estimation performance. Specifically, we proposed a novel framework to train a privacy preserver that converts full-face images into obfuscated counterparts, which are effective for gaze estimation while containing no privacy information. Evaluation on four datasets shows that the obfuscated image can protect users' private information, such as identity and gender, against unauthorized attribute {classification}. Meanwhile, when used directly by the black-box gaze estimator as inputs, the obfuscated images lead to comparable tracking performance to the conventional, unprotected full-face images.  

%% file: 02introduction.tex
\section{Introduction}

Human eye gaze information is pivotal in understanding human attention and the inner workings of the human brain. This unique capability makes the underlying technology, known as gaze tracking, a critical enabler for a wide range of ubiquitous and interaction applications~\cite{bulling2010toward,khamis2018past}. Examples include human activity recognition~\cite{srivastava2018combining,lan2020gazegraph, 9826020,chen2023characteristics}, 
cognitive workload estimation~\cite{duchowski2020low,kosch2018your,fridman2018cognitive}, early detection of autism spectrum disorder~\cite{ 10.1145/3304109.3325818,falck2013eye}, and gaze-based human-computer interaction~\cite{mariakakis2015switchback,wang2020blyncsync,namnakani2023comparing,esteves2015orbits}, among many other applications~\cite{katsini2020role,khamis2018past,bulling2010toward}.

Thanks to the increasing demand for gaze-based applications, many vendors now provide affordable and easy-to-use gaze estimation services~\cite{SeeSo, realeye, pygaze, eyeware, vicarvision, gazerecorder, WeGaze}. By utilizing general-purpose cameras, such as those embedded in tablets~\cite{huang2017tabletgaze}, mobile phones~\cite{Krafka_2016_CVPR}, and public displays~\cite{zhang2015eye}, users can access gaze estimation service either through the cloud server, e.g., RealEye~\cite{realeye} and GazeRecorder~\cite{gazerecorder}, or by installing the software directly on their local devices, e.g., SeeSo~\cite{SeeSo}, WeGaze~\cite{WeGaze}, and EyeWare~\cite{eyeware}. In both cases, users are required to share their full-face images with the service provider, who then takes the images as input to further locate eye regions~\cite{lu2014adaptive, Krafka_2016_CVPR} or directly leverage the whole image \cite{Zhang2020ETHXGaze,zhang2015appearance} for gaze estimation.

Typically, gaze estimation services are managed by commercial entities. This makes the end-to-end system, i.e., the image processing pipeline and the learning-based estimation model, an opaque black box to the user. When querying the gaze estimation services, users do not have any knowledge of how their face images are being processed, stored, or utilized. This problem becomes even more concerning given that %with the reality that 
facial images contain rich information about users' private attributes, such as identity and gender. Thus, when a malicious service provider has access to a large collection of unprotected face images, it can easily infer sensitive user information beyond the intended purpose, posing significant privacy threats to the users~\cite{kroger2020does}.

\begin{figure}[t]
  \centering
  \includegraphics[width=0.82\linewidth]{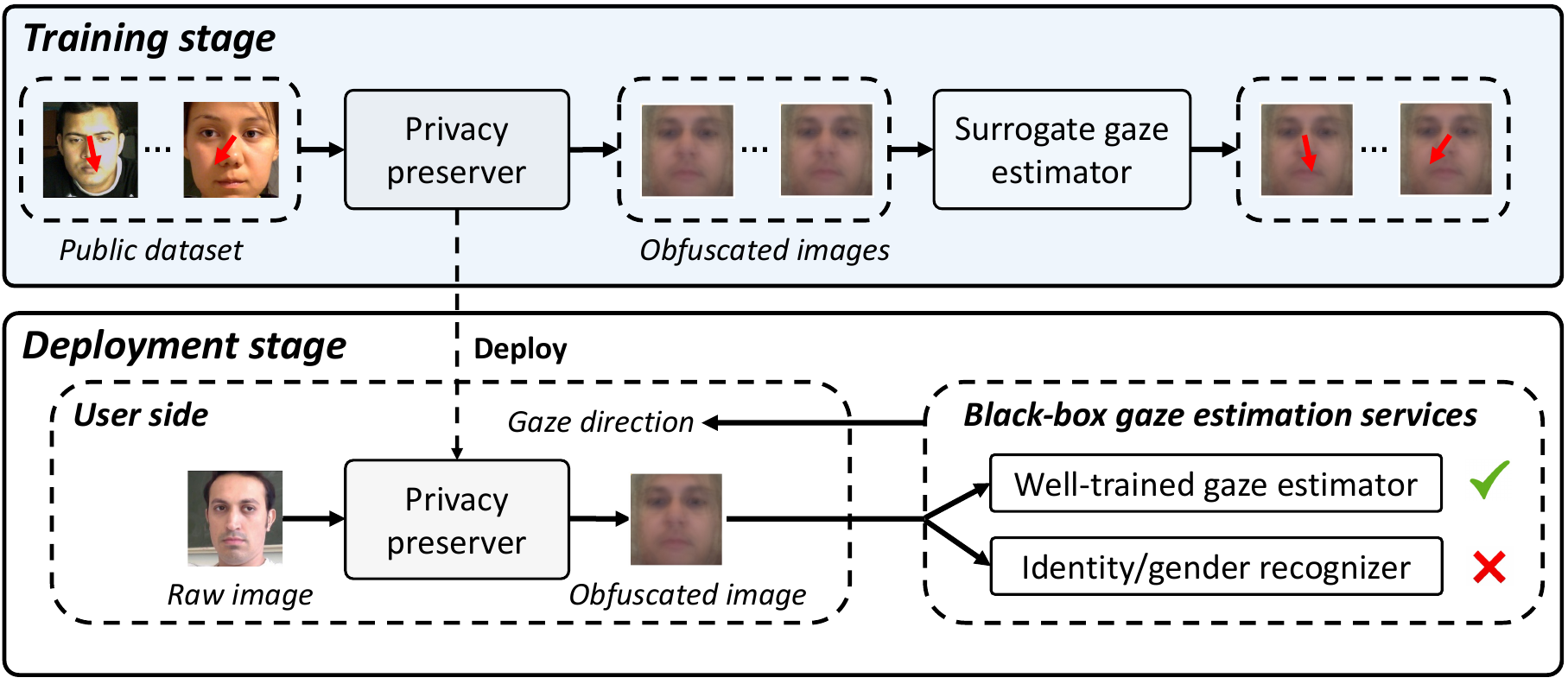}
  \caption{{An illustration of \name}, a framework to preserve users' privacy when they are using black-box gaze estimation services. The core of \name is the privacy preserver, which transforms the original privacy-sensitive full-face image into an obfuscated version as input for the untrusted gaze estimation services. During the training stage, we train the privacy preserver with the assistance of a pre-trained surrogate gaze estimator. After training, the privacy preserve is deployed on the user's device to generate obfuscated images that can be used by the black-box gaze estimation services. This ensures accurate gaze estimation while preventing the user's private attributes, such as gender and identity, from being inferred by the service provider.}
  \label{Fig:teaser_figure}
  % \vspace{-0.1in}
\end{figure}
 
The privacy implications of gaze tracking have started to attract attention from the pervasive computing and eye-tracking communities~\cite{liebling2014privacy,gressel2023privacy}. 
A large body of work~\cite{10.1145/3379157.3390512,eskildsen2021analysis,8998133,david2022your} focuses on the design of obfuscation techniques. These methods aim to eliminate sensitive user information from eye images to prevent iris-based user re-identification without compromising the utility of the estimated gaze data for downstream applications. Although these works have made significant progress in designing privacy-aware gaze-based applications, the question of how to preserve user's privacy in black-box gaze-tracking services remains an open challenge. 

Besides the methods specifically designed for eye tracking, many privacy-preserving techniques have been developed for general-purpose image recognition and detection tasks. A popular solution is to train an encoder-based feature extractor via adversarial learning, so that it can capture task-related features from the original images while eliminating features related to the user's private attributes~\cite{Liu_PAN_2020, Francesco_LPPE_2019,wu2021dapter,xiao2020adversarial}. 
However, these works assume the privacy protector and the user have the full knowledge of the deep learning models utilized by the service provider, which is impractical for the real-world black-box gaze estimation services we considered. Another line of work adds small perturbations in the face images to obstruct unauthorized deep learning models from inferring private user attributes~\cite{oh2017adversarial, Shan_Fawkes_2020}. However, the obfuscated images generated by these methods still contain a substantial amount of private user information that can be visually recognized by human eyes.

In this paper, we propose \name, the first approach that can effectively preserve user's privacy in black-box gaze tracking services without compromising the estimation performance. An overview of \name is shown in Figure \ref{Fig:teaser_figure}. The core component is {the proposed \textit{privacy preserver}}, which operates on the user's side to convert privacy-sensitive full-face images into privacy-enhanced obfuscated images that remain effective for gaze estimation. \name comprises a suite of techniques we developed to tackle two major challenges. 

First, the privacy preserver should eliminate features related to the user's private attributes, i.e., identity and gender, from the original full-face images. To resolve this challenge, {we introduce a novel method that generates an average full-face image from a public dataset and leverages it as a template to transform images of different users}, ensuring that the transformed versions exhibit a similar facial appearance akin to the average full-face image. Second, to maintain good gaze estimation performance, we need to ensure that the essential gaze-related information in the original images is preserved in the obfuscated images. Moreover, the obfuscated images should be readily compatible with the black-box gaze estimator, requiring no additional adaption from the service provider. {To achieve this goal, we train a \textit{surrogate gaze estimator} on a public dataset. As shown in Figure \ref{Fig:teaser_figure}, we leverage the well-trained surrogate gaze estimator to} encourage the privacy preserver to generate obfuscated images that contain features leading to the same gaze direction as the original images. 

In summary, our major contributions are three-fold:
\begin{itemize}
    \setlength\itemsep{0.3em}
    \item We propose \name, the first approach that can effectively preserve users' privacy when using black-box gaze tracking services without compromising the gaze estimation performance.

    \item We propose a novel framework to train a privacy preserver that can be deployed on the user's side to convert private-sensitive full-face images into privacy-enhanced obfuscated counterparts. The obfuscated images are effective for gaze estimation while containing no information about the user's private attributes. 
    
    \item We conduct extensive experiments on four benchmark datasets to demonstrate the effectiveness of \name. The results show that the obfuscated images produced by \name can effectively protect users' private information, such as identity and gender, against unauthorized attribute inference. This protection remains robust even when the malicious attribute recognizer is trained on obfuscated images with correct attribute labels. Moreover, when used directly by the black-box gaze estimator as inputs, the obfuscated images achieve comparable tracking performance to the conventional, unprotected full-face images. Lastly, our system profiling shows that the proposed privacy preserver can be deployed on various computation platforms with low system costs in latency and memory usage.
\end{itemize}

\textbf{Paper roadmap.} The rest of the paper is organized as follows. In Section~\ref{sec:related_work}, we review related work and discuss the research gaps. We then present the detailed design of \name in Section~\ref{sec:method}. Subsequently, we evaluate %the proposed 
\name in Section~\ref{sec:evaluation}, followed by the conclusion in Section~\ref{sec:conlusion}. {The implementation of \name is available at \url{https://github.com/LingyuDu/PrivateGaze}.}

%% file: 03relatedwork.tex
\section{Related Work}
\label{sec:related_work}

\subsection{Gaze Estimation}
Gaze estimation methods can be generally categorized into model-based and appearance-based methods. Model-based methods~\cite{hansen2009eye, guestrin2006general, nakazawa2012point, zhu2007novel} infer gaze directions by constructing geometric models of the eyes from eye images. By contrast, appearance-based methods~\cite{zhang2015appearance, zhang17_cvprw, Krafka_2016_CVPR, tan2002appearance} directly estimate gaze directions from eye images or facial images captured by conventional cameras, such as those embedded in %tablets~\cite{huang2017tabletgaze} or 
mobile devices~\cite{huang2017tabletgaze,Krafka_2016_CVPR}. 

While earlier works in appearance-based methods~\cite{he2015omeg, zhang2015appearance, zhu2007novel} only took eye images as inputs, recent advancements~\cite{Zhang2020ETHXGaze, zhang17_cvprw, Kothari_2021_CVPR, Wang_2022_CVPR, Krafka_2016_CVPR} demonstrate that appearance-based methods can greatly benefit from information contained in facial regions and directly use full-face images for gaze estimation. More recently, the availability of large-scale datasets, such as GazeCapture~\cite{Krafka_2016_CVPR} and ETHXGaze~\cite{Zhang2020ETHXGaze}, combined with advances in deep-learning techniques, have significantly propelled appearance-based gaze estimation forward, enabling more accurate gaze prediction in complex environments with diverse backgrounds and lighting conditions.

The benefits of appearance-based gaze estimation are expected to encourage more users to utilize third-party appearance-based gaze estimation services for developing gaze-based applications, such as user attention estimation~\cite{abdelrahman2019classifying}, cognitive context sensing~\cite{wangwiwattana2018pupilnet,lai2023individualized}, and analyzing user interactions~\cite{modi2023understanding,chong2017detecting}. However, these gaze estimation services often appear as a black box to users for commercial purposes, where users can only utilize the %black-box gaze estimation 
services without access to detailed information about how they operate. %but cannot know details about the services. 
Such black-box gaze estimation services raise %will introduce severe 
significant privacy concerns, as facial images contain rich private information about users. Our work is the first to preserve user privacy when utilizing black-box gaze estimation services.

\subsection{Privacy-Preserving Methods in the Image Domain}
\label{subsec:relatedwork_image_privacy}

There have been several approaches to preserving user privacy in images across various application scenarios. For example, Oh et al.~\cite{oh2017adversarial} employ adversarial image perturbations to raw images to confuse deep learning-based classifiers from recognizing the user's identity. Shan et al.~\cite{Shan_Fawkes_2020} propose adding imperceptible perturbations to images %from users 
before their release, causing identity recognizers trained on these perturbed images to misidentify normal images. However, these perturbed images still retain private attributes and can be easily identified by recognizers trained on the perturbed dataset. In our work, we pursue a stringent privacy objective where attackers cannot infer private attributes from obfuscated images, even when deep learning classifiers have been trained on these obfuscated images with correct attribute labels.

Previous works~\cite{Francesco_LPPE_2019, Liu_PAN_2020} have explored approaches where instead of adding perturbations to raw images, they focus on training an encoder to extract features from raw images %that extracted features contain 
that are useful for utility tasks while excluding information related to private attributes. For instance, Liu et al. \cite{Liu_PAN_2020} utilize adversarial learning to jointly train an encoder, a task-related model, and a private attribute recognizer. The task-related model and the private attribute recognizer are built upon the features extracted by the encoder from raw images. Wu et al. \cite{wu2021dapter} propose DAPter, a method aimed at preserving user privacy during the utilization of task-related inference services on cloud platforms. However, their approach requires access to the task-related model and involves modifying its parameters %of the task-related model 
for effective training, which is impractical for black-box models in real-world scenarios. Our work differs from these approaches by focusing on a more practical scenario where users have no knowledge of the deep learning model employed by the service provider for gaze estimation. 

Face swapping \cite{perov2020deepfacelab, Nirkin_2019_ICCV, Zhu_2021_CVPR, WilsonGazeFaceSwap, Nirkin2018FaceSwap,Cui_2023_CVPR, chen2020simswap, naruniec2020high, Kim_2022_CVPR} and face de-identification \cite{ksameset, jeong2021ficgan, kuang_efficient_face_anony, xue2023face} are potential methods to preserve user privacy by altering the identities of subjects in raw images. However, these techniques may retain other user attributes in the synthesized images \cite{WilsonGazeFaceSwap, naruniec2020high, chen2020simswap, kuang_efficient_face_anony, xue2023face}, such as facial expressions and emotions, which are also privacy-sensitive information that users may wish to protect. %would like to preserve. 
By contrast, our method exclusively maintains gaze features in the synthesized images, offering robust privacy protection for users when utilizing gaze estimation services. Moreover, we observe that synthesized images generated by face swapping from target images of different subjects often exhibit distinct appearances \cite{naruniec2020high, chen2020simswap, Kim_2022_CVPR}, and similarly, de-identified images for different subjects may show visual differences~\cite{jeong2021ficgan}. %while the de-identified images for different subjects also exhibits visual differences \cite{jeong2021ficgan}. 
We believe these inherent properties could be exploited by adversaries to classify users' identities if they can stealthily collect some synthesized images. %of the users. 
In our approach, synthesized images from different subjects maintain similar appearances, and our experiments show that the adversaries cannot correctly classify users' identities even with access to stealthily collected synthesized images.

\subsection{Privacy-preserving Solutions for Eye-tracking Systems}
\label{subsec:relatedwork_eyetracking_privacy}
Privacy-preserving eye tracking has emerged as a significant research topic in recent years due to growing privacy concerns associated with various stages of the eye-tracking pipeline. We categorize existing works %on privacy-preserving eye tracking systems 
\cite{elfares2023federated, steil2019privaceye, 10.1145/3379157.3390512, eskildsen2021analysis, 8998133, 10.1145/3379156.3391364, 9382914, steil2019privacy, 263891} into three groups.

The first group focuses on the data collection stage, aiming to protect users' privacy-sensitive data by preventing its transmission to a central server. For example, Elfares et al. \cite{elfares2023federated} utilize federated learning to train a deep learning-based gaze estimator. This approach retains raw data locally with users to preserve privacy and sends only minimal updates necessary for gaze estimation to the central server. Steil et al. \cite{steil2019privaceye} propose a method to protect the privacy of users and bystanders from scene images captured by a head-mounted eye tracker. Specifically, they develop a method to disable the scene camera upon detecting privacy-sensitive situations and automatically reactivate it when eye movement patterns change.

The second group of works focuses on preserving user privacy in the gaze estimation stage, aiming to mitigate the presence of private attributes in the images used by gaze estimators. For example, John et al. \cite{10.1145/3379157.3390512} introduce pixel-level noise to eye images captured by eye-tracking cameras, effectively thwarting iris authentication attacks. Eskildsen et al. \cite{eskildsen2021analysis} propose various methods to obfuscate eye images, including adding noise and applying non-linear low-pass filters, to prevent identification based on iris patterns. Additionally, John et al. \cite{8998133} propose a hardware-based solution to remove bio-metric information from eye images by inducing optical defocus. This is achieved by increasing the distance between the eyes and the eye-tracking cameras, thereby intentionally blurring the iris region. Bozkir et al. \cite{10.1145/3379156.3391364} train a support vector regression model to estimate gaze direction from synthetic eye images, thereby preserving personal information for users.

{The last group focuses on preserving the private attributes of users contained in the gaze data obtained by gaze estimators.} David-John et al. \cite{9382914} explore various privacy mechanisms such as adding Gaussian noise and temporal downsampling to reduce user identification accuracy based on gaze data features like fixations and saccades. Steil et al. \cite{steil2019privacy} apply differential privacy by adding noise to features extracted from gaze data. They demonstrate that their approach prevents attackers from accurately identifying the user's identity and gender from gaze trajectories, while still maintaining good performance in gaze-based document type recognition tasks. Li et al. \cite{263891} propose a framework that directly applies differential privacy to raw gaze data. Their method can integrate with existing eye-tracking ecosystems and operate in real time, enhancing privacy protection during data processing stage.

In this paper, we focus on addressing privacy concerns during the gaze estimation stage. We propose a novel method to remove private attributes from full-face images while maintaining comparable performance in gaze estimation for black-box gaze estimation services.

%% file: 05method.tex
\section{Method}
\label{sec:method}

In this section, we introduce a novel method, \name, that can convert a normal full-face image into a privacy-perceived full-face image. With the privacy information removed, the privacy-perceived full-face image still contains sufficient information for a black-box gaze estimation service to perform the gaze estimation task. In the following, we first define the threat model to formulate the problem and then detail the design of \name.

\subsection{Threat Model}

\label{subsec:threatmodel}

\noindent\textbf{Black-box gaze estimator.}
With recent developments in gaze estimation, it has become common to include the full-face image of the user as input to the methods~\cite{Krafka_2016_CVPR, zhang17_cvprw}. While existing works in privacy preservation~\cite{wu2021dapter, Liu_PAN_2020, Francesco_LPPE_2019} assume the details of the deep neural network used by the service provider are known, we consider a more practical case where the gaze estimator $\mathcal{G}_b(\cdot)$ is performed by a black-box, deep learning-based model. Specifically, the black-box gaze estimator $\mathcal{G}_b(\cdot)$ \textit{is trained on an unknown dataset $\mathcal{D}_{b}$} that contains raw full-face images and gaze annotations. Users can access gaze estimation services either through the cloud server or by installing the system directly on their local devices. In both cases, the user can only query and request $\mathcal{G}_b(\cdot)$ for service and has no knowledge about its implementation and training details.

\vspace{0.08in}
\noindent\textbf{Privacy concerns.}
The end-to-end gaze estimation system, including the processing pipeline and the deep learning-based gaze estimation model, makes the gaze estimation services untrustworthy. The full-face image of the user can be illegally used for purposes beyond gaze estimation, such as classifying the user's private attributes like identity and gender. This risk persists even if the gaze estimation system is running on local devices, as the gaze estimation services can stealthily share the data collected from the user with the malicious service provider when the local devices are connected to the internet. Therefore, the user would like to remove the private information contained in the full-face images before using them to call the gaze estimation services, without sacrificing gaze estimation performance.

\vspace{0.08in}
\noindent \textbf{Capabilities and goals of the malicious service provider.} 
We assume the malicious service provider can stealthily collect a dataset $\mathcal{D}_p$, comprising images submitted by users for gaze estimation service, along with annotations of private user attributes such as identity and gender. Subsequently, $\mathcal{D}_p$ is used to train classifiers aimed at discerning users' private attributes from images that do not belong to $\mathcal{D}_p$.

\vspace{0.08in}
\noindent\textbf{Our goals.}
In this work, we envision a trustworthy party that provides a privacy preserver $\mathcal{P}(\cdot)$ to protect the user's privacy. As shown in Figure~\ref{Fig:teaser_figure}, during the deployment stage, $\mathcal{P}(\cdot)$ converts the user's original full-face images $x$ into obfuscated images $x'$ that do not contain information related to the user's attributes, such as identity and gender. The user then directly calls the black-box gaze estimator $\mathcal{G}_b(\cdot)$ with the obfuscated image $x'$. Formally, the obfuscated image $x'$ must fulfill the objectives of preserving the user's privacy while ensuring good gaze estimation performance: 
\begin{itemize}
\setlength\itemsep{0.6em}
    \item \textbf{Privacy goal:} The obfuscated image $x'$ cannot be used to correctly classify private attributes of the user, such as identity and gender, even if the malicious service provider trains deep learning-based classifiers on $\mathcal{D}_p$, i.e., a set of $x'$ with accurate labels for these confidential user attributes.
    
    \item \textbf{Utility goal:} The obfuscated image $x'$ can be directly used by $\mathcal{G}_b(\cdot)$ without any adaption needed from the service provider's side. The gaze estimation performance of  $\mathcal{G}_b(\cdot)$ with $x'$ should be similar to the original full-face images.
\end{itemize}

\vspace{0.08in}
\noindent\textbf{Assumption.} 
We assume a public gaze estimation dataset $\mathcal{D}_{w}$ is available, which contains training examples $(x_i,g_i)$, where $x_i$ is the full-face image and $g_i$ is the corresponding gaze annotation. The dataset $\mathcal{D}_{w}$ will be used to train $\mathcal{P}(\cdot)$. Note that $\mathcal{D}_{w}$ is different from $\mathcal{D}_{b}$, as we do not know which dataset has been used by the service provider to train the black-box gaze estimator $\mathcal{G}_b(\cdot)$. 

\begin{figure}[]
  \centering
  \includegraphics[width=0.95\linewidth]{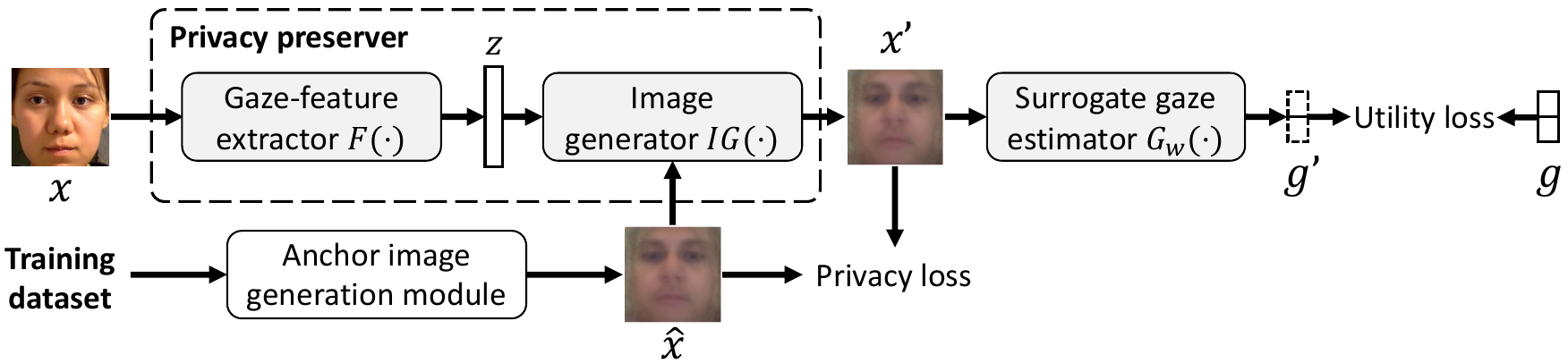}
  \caption{{An overview of \name}, {which comprises the privacy preserver, the anchor image generation module, and the surrogate gaze estimator $\mathcal{G}_w(\cdot)$ trained on the training dataset $\mathcal{D}_{w}$. The privacy preserver includes the gaze-feature extractor $F(\cdot)$ and the image generator $IG(\cdot)$. $F(\cdot)$ extracts gaze features $z$ from the raw images $x$ in the training dataset. $IG(\cdot)$ takes $z$ and a pre-generated image $\hat{x}$ as inputs to form the obfuscated images $x'$. $\hat{x}$ serves as the \emph{anchor image} and is crafted from the training dataset using the proposed anchor image generation module.} Subsequently, we compute the \textit{privacy loss} based on $\hat{x}$ and $x'$ to train $\mathcal{P}(\cdot)$ for the privacy objective. $x'$ is then passed to $\mathcal{G}_w(\cdot)$ to obtain the estimated gaze direction $g'$. Finally, we calculate the \textit{utility loss} based on the gaze annotations $g$ and $g'$ to train the privacy preserver for the utility objective.}
  \label{Fig:Overview}
  % \vspace{-0.1in}
\end{figure}

\subsection{\name}
\label{subsec:privateGaze}

To achieve the design goals, we propose a novel framework \name consisting of a privacy preserver, an anchor image generation module, and the surrogate gaze estimator as shown in Figure \ref{Fig:Overview}. The privacy preserver $\mathcal{P}(\cdot)$ converts unprotected raw images $x$ into obfuscated images $x'$ to protect the private information of users, such as gender and identity contained in $x$. To achieve this goal, the privacy preserver ensures that $x'$, converted from different $x$ (images from different subjects), will exhibit similar facial appearances akin to a pre-generated average full-face image called the anchor image $\hat{x}$. We devise the anchor image generation module (in Section \ref{subsubsec:anchorImageGeneration}) to generate the $\hat{x}$ from the $\mathcal{D}_{w}$.

To achieve the utility goal, the privacy preserver $\mathcal{P}(\cdot)$ is designed to extract gaze features $z$ from $x$ and generate $x'$ that maintains these features for effective gaze estimation. Specifically, $\mathcal{P}(\cdot)$ consists of the gaze-feature extractor $F(\cdot)$ and the image generator $IG(\cdot)$. $F(\cdot)$ extracts gaze features $z$ from the input $x$ (detailed in Section \ref{subsubsec:gazeFeatureExtractor}). $IG(\cdot)$ takes $z$ along with $\hat{x}$ as inputs to generate the privacy-preserved $x'$ (detailed in Section \ref{subsubsec:imageGenerator}). The generated $x'$ has a similar appearance to $\hat{x}$ while preserving the gaze-related information $z$ from $x$ for accurate gaze estimation. To train $\mathcal{P}(\cdot)$, we construct a surrogate gaze estimator $\mathcal{G}_w(\cdot)$ trained on the $\mathcal{D}_{w}$, which performs the gaze estimation training with input $x'$. In this way, we are able to maximize the information in $x'$ for the gaze estimation task. 

\subsubsection{Anchor image generation module.}
\label{subsubsec:anchorImageGeneration}
Below, we present a novel method for generating the anchor image $\hat{x}$ from a public dataset. The anchor image serves as a template for the obfuscated images $x'$, ensuring they exhibit a facial appearance similar to $\hat{x}$. This allows us to manipulate the appearances of $x'$ to preserve user's privacy while achieving the utility goal. 

A major challenge in achieving this utility goal is training $\mathcal{P}(\cdot)$ with the surrogate gaze estimator $\mathcal{G}_w(\cdot)$, while aiming for good gaze estimation performance on the black-box gaze estimator $\mathcal{G}_b(\cdot)$. To address this challenge, we carefully generate the anchor image $\hat{x}$ to ensure that both $\mathcal{G}_w(\cdot)$ and $\mathcal{G}_b(\cdot)$ yield similar gaze estimation results on $\hat{x}$. This strategy enables $\mathcal{G}_w(\cdot)$ and $\mathcal{G}_b(\cdot)$ to achieve comparable gaze estimation performance on the obfuscated images $x'$, as they share similar appearances with the anchor image. 

In detail, $\hat{x}$ is an average full-face image created by blending facial images selected from the training dataset. The design ensures that $x'$ synthesized from $\hat{x}$ does not closely resemble any individual subject. {Then, we design the method to obtain $\hat{x}$ that can lead to similar gaze estimation results from $\mathcal{G}_w(\cdot)$ and $\mathcal{G}_b(\cdot)$.} Specifically, we first randomly sample $N$ raw images $\{x_i\}_{i=1}^N$ from the $\mathcal{D}_{w}$ and use them to query both $\mathcal{G}_w(\cdot)$ and $\mathcal{G}_b(\cdot)$, where the two gaze estimators return the corresponding gaze estimation results $\{\mathcal{G}_w(x_i)\}_{i=1}^N$ and $\{\mathcal{G}_b(x_i)\}_{i=1}^N$, respectively. We then calculate the $L1$ norm between the gaze directions estimated by $\mathcal{G}_w(\cdot)$ and $\mathcal{G}_b(\cdot)$ for each image $x_i$ to obtain the list $\{D_g(x_i)\}_{i=1}^N$, where $D_g(x_i)=|\mathcal{G}_w(x_i)-\mathcal{G}_b(x_i)|_1$. 
After that, we sort the list $\{D_g(x_i)\}_{i=1}^N$ in the ascending order. 
We use $\{x_k\}_{k=1}^N$ to denote the set of raw images after sorting, which satisfies $D_g(x_k)\leq D_g(x_{k+1})$. We use $Ave(m)$ to denote the average full-face image calculated from the first $m$ images in $\{x_k\}_{k=1}^N$ by $Ave(m)=\frac{1}{m}\sum_{k=1}^{m}x_k$. 
We generate $M$ average full-face images by varying $m$ from $K_1+1$ to $K_1+M$, where $K_1>1$ and $K_1+M\leq N$. 
Finally, we query $\mathcal{G}_w(\cdot)$ and $\mathcal{G}_b(\cdot)$ with the $M$ average full-face images and select the one that leads to the minimum $L_1$ norm between the outputs of $\mathcal{G}_w(\cdot)$ and $\mathcal{G}_b(\cdot)$ as the anchor image $\hat{x}$. 
We set $K_1=15$, $M=35$, and $N=500$. {The number of queries for $G_b(\cdot)$ is determined by $N$. The parameters $K$ and $M$ determine the minimum and maximum number of full-face images utilized in generating the anchor image, respectively. A higher value of $K$ ensures the anchor image to be distinct from any single subject, whereas a larger $M$ may include undesired full-face images, potentially resulting in significantly different gaze estimation results for $\mathcal{G}_w(\cdot)$ and $\mathcal{G}_b(\cdot)$ during the anchor image generation process.}
We show the $\hat{x}$ obtained from GazeCapture dataset \cite{Krafka_2016_CVPR} in Figure~\ref{Fig:PrivacyPreserver} and summarize the procedure of generating the anchor image in Algorithm \ref{Alg:anchorImage}.

\begin{algorithm}[t]
\caption{Anchor image generation}\label{Alg:anchorImage}
\begin{algorithmic}[1]
\Require $\mathcal{D}_{w}$, $\mathcal{G}_b(\cdot)$, $\mathcal{G}_w(\cdot)$, $K_1=15$, $m= 1$, $M = 35$, $N = 500$, and a list of candidate anchor image $Anc=\{\}$.
\State Randomly sample $N$ images from $\mathcal{D}_w$ to form $\{x_i\}_{i=1}^N$;
\State Query both 
$\mathcal{G}_{b}(\cdot)$ and $\mathcal{G}_w(\cdot)$ with $\{x_i\}_{i=1}^N$ to obtain the list $\{D_g(x_i)\}_{i=1}^N$, where $D_g(x_i)=|\mathcal{G}_w(x_i)-\mathcal{G}_b(x_i)|_1$;
%\State Obtain the list $\{D_g(x_i)\}_{i=1}^N$, where $D_g(x_i)=|\mathcal{G}_w(x_i)-\mathcal{G}_b(x_i)|_1$;
\State Sort $\{D_g(x_i)\}_{i=1}^N$ in the ascending order;
\State Obtain the set of raw images after sorting $\{x_k\}_{k=1}^N$, where $D_g(x_k)\leq D_g(x_{k+1})$;
\While{$K_1+m\leq M$}
    \State Calculate the average full-face image $Ave(m)=\frac{1}{m}\sum_{k=1}^{m}x_k$;
    \State $Anc.$append$(Ave(m))$;
    \State $m \gets m+1$;
\EndWhile
\State Query $\mathcal{G}_{b}(\cdot)$ and $\mathcal{G}_w(\cdot)$ with $Anc=\{Ave(m)\}_{m=1}^{M}$ to obtain the list $\{D_g(Ave(m))\}_{m=1}^M$;%, where $D_g(Ave(m))=|\mathcal{G}_w(Ave(m))-\mathcal{G}_b(Ave(m))|_1$
\State $\hat{x}=Ave(\hat{m})$, where $\hat{m}=\arg\min_{m} \{D_g(Ave(m))\}_{m=1}^M$;
\Ensure The anchor image $\hat{x}$.
\end{algorithmic}
\end{algorithm}

\subsubsection{Gaze-feature extractor.}
\label{subsubsec:gazeFeatureExtractor}
\begin{figure}[]
  \centering
  \includegraphics[width=0.58\linewidth]{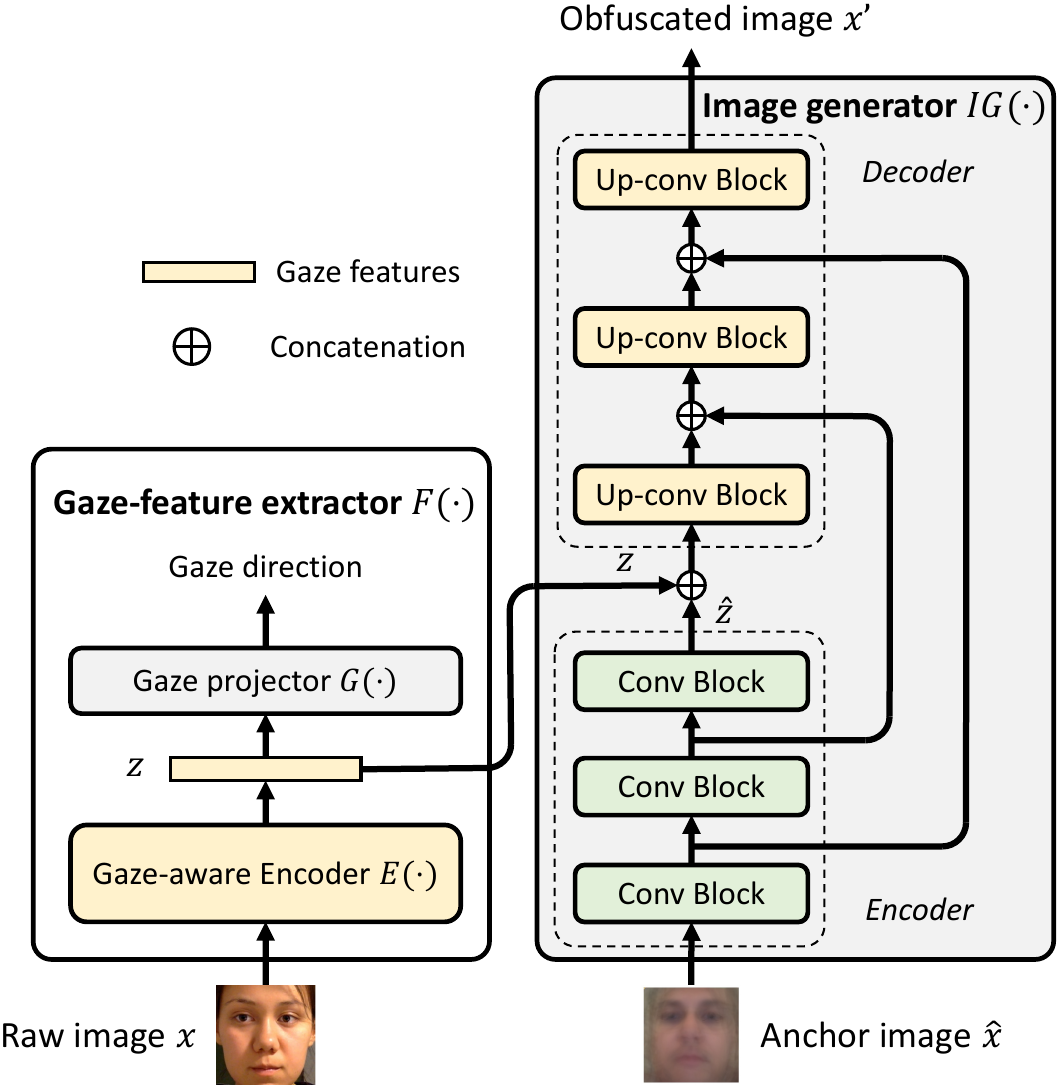}
  \caption{{The overall design of the privacy preserver $\mathcal{P}(\cdot)$}, which consists of the gaze-feature extractor $F(\cdot)$ and the image generator $IG(\cdot)$. $F(\cdot)$ extracts gaze features $z$ from the raw image $x$ of the user. $IG(\cdot)$ takes the extracted gaze features $z$ and the anchor image $\hat{x}$ as inputs to generate the privacy-preserved obfuscated image $x'$. $x'$ has a similar appearance to $\hat{x}$ while retaining the gaze features extracted from $x$. Only the components with color-coded yellow will be deployed on the user's device after training for privacy preservation.}
  \label{Fig:PrivacyPreserver}
  % \vspace{-0.1in}
\end{figure}

To ensure the obfuscated images $x'$ are effective for gaze estimation, $\mathcal{P}(\cdot)$ extracts the gaze features $z$ from $x$ using the gaze-feature extractor $F(\cdot)$. As shown in the left part of Figure~\ref{Fig:PrivacyPreserver}, $F(\cdot)$ comprises the gaze-aware encoder $E(\cdot)$ and the gaze projector $G(\cdot)$. Specifically, $E(\cdot)$ takes $x$ as input and outputs a feature map $z$. To encourage $E(\cdot)$ in capturing the most essential and meaningful gaze-related features, $z$ is fed into a nonlinear gaze projector $G(\cdot)$ to estimate gaze direction. We define the \textit{gaze estimation loss} in training $E(\cdot)$ and $G(\cdot)$ as follow:
\begin{equation}%\large
    \mathcal{L}_g=\sum_{(x_i,g_i)\in\mathcal{D}_w}\ell(G(E(x_i)),g_i),
    \label{Eq:gaze_mlp}
\end{equation}
where $\ell(\cdot)$ is the $L_1$ loss function. The $G(\cdot)$ will be discarded in the deployment stage, and only the gaze features $z$ are sent to the image generator $IG(\cdot)$.

\subsubsection{Image generator.} 
\label{subsubsec:imageGenerator}
The image generator $IG(\cdot)$ is designed to synthesize the obfuscated images $x'$ from gaze features $z$ such that $x'$ can be effectively used for gaze estimation by $\mathcal{G}_b(\cdot)$ while not containing private attributes from $z$. To achieve this goal, $IG(\cdot)$ generates $x'$ with appearances similar to the anchor image $\hat{x}$ while preserving the gaze features extracted from $x$. Specifically, $IG(\cdot)$ takes both $z$ and $\hat{x}$ as inputs to generate $x'$.

The structure of $IG(\cdot)$ is depicted in the right part of Figure~\ref{Fig:PrivacyPreserver}, which adopts an encoder-decoder architecture. The encoder comprises several convolutional blocks (Conv Block) and takes $\hat{x}$ as input to produce a feature map $\hat{z}$ that has the same spatial dimension as the gaze features $z$ extracted by the gaze-aware encoder $E(\cdot)$. $\hat{z}$ and $z$ are concatenated and then fed into the decoder. 
Similar to the encoder, the decoder comprises several up-convolutional blocks (Up-conv Block). Each up-convolutional block involves upsampling the feature map followed by several convolutional layers. In our current design, each convolutional block consists of two convolutional layers, and each up-convolutional block has three convolutional layers. As shown in Figure \ref{Fig:PrivacyPreserver}, to ensure that $x'$ can closely resemble $\hat{x}$, we concatenate the output of each up-convolutional block with the corresponding feature map from the encoder. The resulting combined feature map is then used as the input for the next up-convolutional block in the sequence. 

We denote the process of generating $x'$ from $z$ and $\hat{x}$ as $x'=IG(\{z,\hat{x}\})$. Note that in Figure~\ref{Fig:PrivacyPreserver}, we use three convolutional blocks and three up-convolutional blocks to illustrate the structure of $IG(\cdot)$. However, in practice, the number of these convolutional and up-convolutional blocks can vary based on design considerations and we are using four convolutional blocks and four up-convolutional blocks in our current design.

\subsection{Training of the Privacy Preserver}

First, to achieve the privacy objective, it is crucial that $x'$, generated from different raw images, maintains a uniform appearance similar to $\hat{x}$. Therefore, we define the privacy loss as follows:
\begin{equation}
    \mathcal{L}_{privacy} = \sum_{(x_i,g_i)\in\mathcal{D}_{w}}1-\text{MS-SSIM}\left(\hat{x}, IG\left(\{E(x_i),\hat{x}\}\right)\right),
\end{equation}
where $\text{MS-SSIM}(\cdot,\cdot)$ is a function that calculates the multi-scale structural similarity~\cite{1292216}, measuring the similarity between two images with values in the range of $[0,1]$. A larger value of $\text{MS-SSIM}$ indicates greater similarity between the two images. 
% \name terms $\mathcal{L}_{privacy}$ as the privacy loss. 

% \subsubsection{Utility loss.} 
Second, to achieve the utility objective, we utilize $x'$ as input to $\mathcal{G}_{w}(\cdot)$ to obtain the estimated gaze direction $g'$ for $x'$, and then train $\mathcal{P}(\cdot)$ to ensure $g'$ closely approximates $g$. {$\mathcal{G}_w(\cdot)$ is trained on raw images, and its parameters are frozen during the training of $\mathcal{P}(\cdot).$}
Besides, the gaze estimation loss $\mathcal{L}_g$, defined in Equation \ref{Eq:gaze_mlp}, encourages the extraction of gaze features from raw images, thereby contributing to the utility objective. For training the privacy preserver to achieve the utility objective, we define the following utility loss:
\begin{equation}
    \mathcal{L}_{utility}=\sum_{(x_i,g_i)\in\mathcal{D}_{w}}\ell\left(\mathcal{G}_{w}\left(IG\left(\{E(x_i),\hat{x}\}\right)\right),g_i\right) + \mathcal{L}_g, 
    \label{Eq:utility_white}
\end{equation}
where $\ell(\cdot)$ is the $L_1$ loss function. The first term of $\mathcal{L}_{utility}$ aims to minimize the $L_1$ norm between $\mathcal{G}_{w}(x')$ and $g$. The second term encourages the extraction of gaze features to generate $x'$.

Putting them all together, the final optimization objective for training $\mathcal{P}(\cdot)$ is the weighted sum of $\mathcal{L}_{utility}$ and $\mathcal{L}_{privacy}$:
\begin{equation}
\mathcal{L} = \mathcal{L}_{utility} + \lambda\mathcal{L}_{privacy}, 
\label{Eq:loss_function}
\end{equation}
where $\lambda$ is the weight that balances the trade-off between the utility and privacy objectives. \name optimizes $\mathcal{P}(\cdot)$ by minimizing $\mathcal{L}$. During training, \name samples a minibatch from $\mathcal{D}_{w}$ to calculate $\mathcal{L}$, and trains $\mathcal{P}(\cdot)$ by gradient descent. We summarize the training procedure for $\mathcal{P}(\cdot)$ in Algorithm \ref{Alg:PrivateGazeTraining}.

\begin{algorithm}[t]
\caption{Training algorithm of the privacy preserver}\label{Alg:PrivateGazeTraining}
\begin{algorithmic}[1]
\Require $\mathcal{D}_{w}$, $\mathcal{G}_w(\cdot)$, and $\hat{x}$ generated by Algorithm \ref{Alg:anchorImage}.
\State Randomly initialize the parameters $\psi$ of $\mathcal{P}(\cdot)$;
\State \textbf{for} each training step \textbf{do}
\State \quad Sample the minibatch $B$ of $N_b$ images from $\mathcal{D}_{w}$;
\State \quad Calculate the privacy loss over $B$ by $\mathcal{L}_{privacy} = \frac{1}{N_b}\sum_{(x_i,g_i)\in B}1-\text{MS-SSIM}\left(\hat{x},IG\left(\{E(x_i),\hat{x}\}\right)\right)$;
\State \quad Calculate the utility loss over $B$ by $\mathcal{L}_{utility} = \frac{1}{N_b}\sum_{(x_i,g_i)\in B}\ell\left(\mathcal{G}_{w}\left(IG\left(\{E(x_i),\hat{x}\}\right)\right),g_i\right) + \mathcal{L}_g$;
\State \quad Update $\psi$ by gradient descent: minimize $\mathcal{L}_{utility} + \lambda\mathcal{L}_{privacy}$;
\Ensure $\mathcal{P}(\cdot)$ with trained parameters $\psi$.
\end{algorithmic}
\end{algorithm}

\subsection{Deployment of the Privacy Preserver}
In the deployment stage, the raw images collected from users are initially transformed into obfuscated images using the trained privacy preserver. Users then use these obfuscated images when calling the black-box gaze estimation services to protect their privacy. To reduce the computational cost, only specific components of the privacy preserver (highlighted in yellow in Figure~\ref{Fig:PrivacyPreserver}) need to be deployed on the user's device. Specifically, for the gaze-feature extractor, only the gaze-aware encoder needs to be deployed, as the image generator exclusively utilizes gaze features from this encoder. For the image generator, because the anchor image remains consistent for different raw images, the encoder can be omitted. Instead, the feature maps generated by each convolutional block of the encoder are preserved as \emph{appearance features}. Consequently, deployment requires only the decoder of the image generator and the appearance features to generate obfuscated images.

By deploying only these essential components, computational resources are optimized while achieving the system goal. The evaluation section includes a latency measurement of the privacy preserver, demonstrating its suitability for deployment across various hardware platforms without introducing much computational latency.

%% file: 06evaluation.tex
\section{Evaluation}
\label{sec:evaluation}
In this section, we conduct a comprehensive evaluation of \name. We first introduce the datasets, followed by the methods for comparison and evaluation metrics. We then present the evaluation results on privacy and utility objectives. Next, we conduct ablation studies to investigate the impact of different design choices on the performance of \name. Finally, we evaluate system performance, measuring the processing time and memory usage when deploying the proposed privacy preserver on various hardware platforms. 

\subsection{Datasets}
We consider the following four public gaze estimation datasets in our evaluation:

\vspace{0.08in}
\noindent\textbf{ETHXGaze} \cite{Zhang2020ETHXGaze} is a comprehensive dataset collected from 110 subjects in a laboratory environment, showcasing a wide range of head positions, lighting conditions, and individual appearances. It includes one training set and two testing sets. Our evaluation only utilizes the training set, which contains 80 subjects, as it is the only subset with gaze annotations. The images in this set have a resolution of $224\times 224$. 

\vspace{0.08in}
\noindent\textbf{GazeCapture} \cite{Krafka_2016_CVPR} is a large-scale dataset collected from over 1,450 individuals in real-world environments. It includes nearly 2.5 million images taken with the front-facing cameras of smartphones, displaying a wide array of lighting conditions, head poses, user appearances, and backgrounds. In our preprocessing of this dataset, we adopted the method outlined in \cite{Zheng2020NeurIPS} to normalize the facial images, initially bringing them to a resolution of $128\times 128$. Subsequently, we resized these images to a resolution of $224\times 224$.

\vspace{0.08in}
\noindent\textbf{MPIIFaceGaze} \cite{zhang17_cvprw} is a dataset of full-face images from 15 subjects, {including nine males and six females.} The images were captured during the participants' routine laptop usage. It includes 3,000 images per subject, featuring a diverse range of head positions, lighting environments, and backgrounds. For our purposes, we utilize the normalized version of the dataset, as released by the authors. This normalized dataset maintains an image resolution of $224\times 224$. 

\vspace{0.08in}
\noindent\textbf{ColumbiaGaze} \cite{CAVE_0324} was gathered in a controlled laboratory setting, and features data from 56 subjects, including 32 males and 24 females. It is unique for its structured capture of five distinct head poses for each subject. The dataset comprises a total of 105 images per subject, representing combinations of three vertical and seven horizontal gaze directions. In our processing of this dataset, we extract the facial region from the original images and subsequently resize these cropped patches to a uniform resolution of $224\times 224$.
% \end{itemize}
\vspace{0.08in}

\begin{figure}[]
  \centering
  \includegraphics[width=0.995\linewidth]{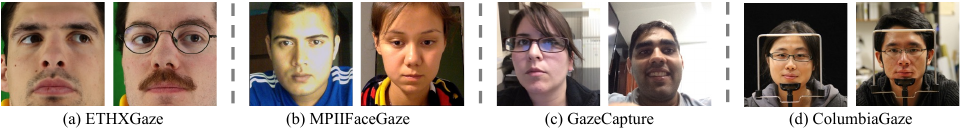}
  \caption{{Illustration of images sampled from the four gaze estimation datasets.} Our selection of datasets covers a broad spectrum of mobile gaze tracking scenarios: from smartphone usage (GazeCapture) to laptop use cases (MPIIFaceGaze), and to ubiquitous web cameras (ETHXGaze and ColumbiaGazze) that widely appear in many daily devices.}
  \label{Fig:DatasetIllustration}
  \vspace{-0.10in}
\end{figure}

Figure~\ref{Fig:DatasetIllustration} showcases full-face images sampled from the four datasets under discussion. These images represent a wide range of scenarios in mobile gaze tracking, from the use of front-facing smartphone cameras (GazeCapture) to laptop-based interactions (MPIIFaceGaze). Moreover, the ETHXGaze and ColumbiaGaze datasets represent scenarios where web cameras, commonly found in a variety of ubiquitous devices, are used for gaze tracking.

\subsection{Comparison Methods}
\label{subsec:comparedMethods}
We compare \name with the following {six} methods:

\vspace{0.06in}
\noindent\textbf{Targeted Projected Gradient Descent Attack (TPGD):} 
In this method, we first feed the original image $x_i$ from the testing dataset to the surrogate gaze estimator $\mathcal{G}_w(\cdot)$ to output the targeted gaze direction $\overline{g}_i$. We then implement the targeted projected gradient descent attack  (TPGD)~\cite{madry2017towards} to create perturbations $\Delta(x_i)$ for the anchor image $\hat{x}$ so that $\mathcal{G}_w(\cdot)$ outputs $\overline{g}_i$ for $\hat{x}+\Delta(x_i)$. This self-created baseline falls under the same category as the proposed \name as the targeted attack~\cite{Dong_2018_CVPR, Feng_2023_CVPR, pmlr-v80-ilyas18a}, where the goal of \name is to make the surrogate gaze estimator $\mathcal{G}_w(\cdot)$ output a targeted gaze annotation $g_i$. Finally, we take $\hat{x}+\Delta(x_i)$ as the obfuscated image and feed it to the black-box gaze estimator $\mathcal{G}_b(\cdot)$.

\vspace{0.06in}
\noindent\textbf{Gaussian Differential Privacy (GauDP):} {The local model of differential privacy~\cite{dwork2014algorithmic} is a potential way to preserve the privacy of user when the central server is not trusted. In this baseline method, we introduce Gaussian noise to each color component $x[p,q,t]$ of every pixel in the raw image $x$. Specifically, we define the mechanism operating on $x[p,q,t]$ as $M(x[p,q,t])=x[p,q,t]+\xi$, where $\xi \sim \mathcal{N}(0, Sen^2 /\epsilon^2)$, with $Sen$ representing the sensitivity of the color component's value, and $\epsilon$ denoting the privacy parameter. Since the value of each color component in the image ranges between 0 and 1, $Sen$ is set to 1. As demonstrated in \cite{dong2022gaussian}, $M$ satisfies $\epsilon$-Gaussian Differential Privacy. In our evaluation, we explore GauDP with $\epsilon$ ranging from $0.1$ to $0.5$.}

\vspace{0.06in}
\noindent\textbf{Image Pixelization with Differential Privacy (IP-DP):} %We implement 
IP-DP \cite{fan2018image} %as a baseline for comparison, which 
is the state-of-the-art differential privacy method operating on images. Specifically, IP-DP first performs pixelization on the raw images, then adds noise sampled from a Laplace distribution to the pixelized images. The Laplace distribution has a mean of 0 and a scale of $\Delta P_b/\epsilon$, where $\Delta P_b$ is the global sensitivity of the pixelized images and $\epsilon$ is the privacy parameter. We evaluate IP-DP with different values of $\epsilon$, including $0.3$, $3.0$, and $5.0$.

\vspace{0.06in}
\noindent\textbf{Feature-space Differential Privacy (FS-DP):} A variety of state-of-the-art differential privacy techniques \cite{xue2021dp, croft2022differentially, wen2022identitydp, li2021differentially} add perturbations to features extracted by an encoder from raw images, rather than directly modifying the raw images themselves. We adapt these techniques for gaze estimation by training a variational auto-encoder (VAE) \cite{kingma2013auto} with the surrogate gaze estimator $\mathcal{G}_w(\cdot)$. 
During training, the VAE takes raw images $x$ as inputs and outputs reconstructed images $\hat{x}$, which are then processed by $\mathcal{G}_w(\cdot)$ to obtain gaze directions $g'$. In addition to the original VAE loss function, we introduce an extra loss term to optimize for utility, aiming to minimize the $L_1$ norm between $g'$ and the gaze annotation $g$. In the deployment stage, the encoder of the VAE extracts $d$ dimensional features $f$ from $x$. We define the mechanism operating on $f$ as $M(f)=f+\xi$, where $\xi$ is sampled from a Laplace distribution $Lap(\Delta f/(d\epsilon))$. Here, $\Delta f$ represents the sensitivity of the features and $\epsilon$ denotes the privacy parameter. The perturbated features are then fed into the decoder of the VAE to generate the obfuscated images. Following Xue et al.~\cite{xue2021dp}, we calculate the sensitivity as $\Delta f = \max_{f_i,f_j}\|f_i-f_j\|_1$, where $f_i$ and $f_j$ are features extracted from different raw images. We examine FS-DP with $\epsilon$ ranges from $1.0$ to $3.0$.

\vspace{0.06in}
\noindent\textbf{B-DAP:} This method is adapted from DAPter \cite{wu2021dapter}, which aims to preserve user privacy in a white-box setting where the user possesses full knowledge of the deep learning model used by the service provider. The original DAPter employs a generative model-based image converter to generate obfuscated images. The image converter is trained to minimize an entropy reduction loss for privacy preservation and a task-related loss defined on the outputs of the target model for the utility objective. We adapt this method by using the surrogate gaze estimator $\mathcal{G}_w(\cdot)$ as the target model during the training stage. Specifically, the adapted method B-DAP trains the image converter by minimizing both the entropy reduction loss and the $L_1$ loss between $g'_i$ and the annotation $g_i$. During testing, we evaluate the effectiveness of B-DAP using the black-box gaze estimator $\mathcal{G}_b(\cdot)$.

\vspace{0.06in}
\noindent\textbf{MaxP:} A straightforward baseline is to minimize the similarity between the obfuscated and the raw image while ensuring the surrogate gaze estimator can still perform gaze estimation with the obfuscated image. Specifically, we train an auto-encoder that takes the raw image $x_i$ as input and outputs the obfuscated image $x'_i$. During training, we feed $x'_i$ to the surrogate gaze estimator to obtain the estimated gaze direction $g'_i$. The auto-encoder is trained to minimize the similarity between $x_i$ and $x'_i$ to achieve the privacy objective, and to minimize the $L_1$ loss between $g'_i$ and the annotation $g_i$ for the utility objective. In the testing stage, we feed the obfuscated image to the black-box gaze estimator $\mathcal{G}_b(\cdot)$.

\subsection{Evaluation Setup and Metrics}
\label{subsec:evaluationSetup}

In the evaluation, we designate ETHXGaze as the unknown dataset $\mathcal{D}_{b}$ to train the black-box gaze estimator $\mathcal{G}_b(\cdot)$, as it contains the most diverse head poses and gaze variations. {Note that, $\mathcal{G}_b(\cdot)$ is trained on the raw images of $\mathcal{D}_{b}$.} We use GazeCapture to train both the privacy preserver $\mathcal{P}(\cdot)$ and the surrogate gaze estimator $\mathcal{G}_w(\cdot)$, {with 80\% of the images used for training and 20\% for validation.} We select MPIIFaceGaze and ColumbiaGaze to evaluate the performance of $\mathcal{P}(\cdot)$. Our primary goal is to preserve the private attributes, i.e., identity and gender, of the individuals in these datasets while ensuring that gaze estimation performance remains robust when using the black-box gaze estimator. 

For utility measurement, we measure the gaze estimation error, i.e., the average angular error, via the black-box gaze estimator $\mathcal{G}_b(\cdot)$. We take the obfuscated images generated by our \name and other baselines as the inputs to assess the utility performance. We conduct this evaluation on both MPIIFaceGaze and ColumbiaGaze datasets. To evaluate the performance in privacy protection, we focus on preserving gender and identity as the two user attributes. {For both the MPIIFaceGaze and ColumbiaGaze datasets, we start by randomly selecting $80\%$ of the images that have accurate identity and gender labels. We then apply each of the seven methods under examination, i.e., PrivateGaze, GauDP, IP-DP, FS-DP, MaxP, B-DAP, and TPGD, to generate obfuscated images and form seven corresponding $\mathcal{D}_p$ sets. The $\mathcal{D}_p$ is used to train an identity recognizer and a gender recognizer. Lastly, we apply each of the seven methods on the remaining $20\%$ of images to create corresponding testing set comprising obfuscated images. We report the recognition accuracy on this testing set for performance evaluation.}

\subsection{Implementation}
To demonstrate the generalization of \name, we employed various neural network architectures to construct the black-box gaze estimator $\mathcal{G}_b(\cdot)$ including ResNet18 \cite{He_2016_CVPR}, MobileNetV2 \cite{Sandler_2018_CVPR}, ShuffleNet \cite{Zhang_2018_CVPR}, VGG11 \cite{simonyan2014very}, and EfficientNet \cite{pmlr-v97-tan19a}. Since \name and TPGD generate different obfuscated images for each black-box gaze estimator, the reported results are averaged across these five architectures. By contrast, the obfuscated images generated using GauDP, IP-DP, FS-DP, MaxP, and B-DAP are independent of the black-box gaze estimator used. 
The surrogate gaze estimator $\mathcal{G}_w(\cdot)$ is implemented using the ResNet18 architecture~\cite{He_2016_CVPR}. The classifiers used for identity and gender recognition are implemented using ResNet18. The encoder of the image generator $IG(\cdot)$ shares the same structure as $E(\cdot)$, ensuring that features extracted from the raw image $x_i$ and the anchor image $\hat{x}$ have identical spatial dimensions. The decoder of the image generator consists of four up-convolutional blocks, each containing an upsampling of feature maps followed by three convolutional layers.

We develop \name using the PyTorch framework and use the Adam optimizer~\cite{kingma2014adam}. The standard learning rate is set to $0.001$, unless specified otherwise. The privacy preserver is trained over 12,000 steps with a mini-batch size of 25 for all evaluation scenarios. The classifiers for identity and gender recognition are trained for 20 and 5 epochs, respectively. The surrogate gaze estimators, trained on the GazeCapture dataset, undergo 5 epochs of training. The black-box gaze estimators, using different structures, are trained for 25 epochs. The learning rates for training both the surrogate and black-box gaze estimators are set to $0.0001$. We fix the value of $\lambda$ in Equation \ref{Eq:loss_function} at 75 and conduct ablation studies to assess the impact of $\lambda$ on the performance of \name.

\subsection{Performance in the Privacy Goal}

\label{subsec:PrivacyPerformance}
The evaluation results for the effectiveness of various privacy-preserving methods are summarized in Table \ref{Tab:privacy}. {For \name, it is important to note that different anchor images are generated when the black-box estimator follows different network architectures. This variation occurs because the image generation module queries $\mathcal{G}_b(\cdot)$ to generate the anchor image $\hat{x}$. 
We report the averaged identity recognition accuracy and gender recognition accuracy for \name over these different structures of $\mathcal{G}_b(\cdot)$.}

From Table \ref{Tab:privacy}, it is evident that the average identity and gender recognition accuracies on images obfuscated by \name are notably low, with values of 6.3\% and 1.1\% respectively, and minimal standard deviations of 0.2 and 0.1 across different structures of $\mathcal{G}_b(\cdot)$. 
Additionally, the average gender recognition accuracies are 62.0\% for the MPIIFaceGaze dataset and 53.9\% for the ColumbiaGaze dataset, both with a standard deviation of 0.0 across different structures of $\mathcal{G}_b(\cdot)$. 
The results underscore the effectiveness of \name in significantly reducing the recognizability of identity and gender attributes in the obfuscated images. The attacker cannot train an effective identity recognizer or gender recognizer on the obfuscated images generated by \name, even when they have access to correct identity and gender labels of the obfuscated images. The minimal standard deviations indicate that \name achieves consistent privacy performance across various architectures of $\mathcal{G}_b(\cdot)$, demonstrating its robustness and generalizability. 

\begin{table}
    \centering
    \caption{Identity and gender recognition accuracies (\%) on obfuscated images generated by \name and other baseline methods evaluated on (a) MPIIFaceGaze and (b) ColumbiaGaze datasets. We report the results of GauDP, IP-DP, and FS-DP with different $\epsilon$ values. The proposed \name can effectively preserve the private attributes against the attackers who train their classifiers on obfuscated images annotated with correct identity and gender labels. Lower recognition accuracy indicates better performance in privacy protection. ``W/o Defense'' denotes the scenario where unprotected original images are used for gaze estimation services, with attribute classifiers trained and tested on the original images.}
    \vspace{-0.1in}
    \label{Tab:privacy}
    \begin{subtable}{}
        \centering
        \resizebox{\textwidth}{!}{%
\begin{tabular}{ccccccccccccccccc}
\Xhline{2\arrayrulewidth}
\multirow{2}{*}{Attributes} & \multirow{2}{*}{{w/o Defense}} & \multicolumn{3}{c}{{GauDP}} &  & \multicolumn{3}{c}{{IP-DP}} &  & \multicolumn{3}{c}{{FS-DP}} & \multirow{2}{*}{MaxP} & \multirow{2}{*}{BDAP} & \multirow{2}{*}{TPGD} & \multirow{2}{*}{PrivateGaze} \\ \cline{3-5} \cline{7-9} \cline{11-13}
                            &                              & {0.1}     & {0.3}    & {0.5}    &  & {0.3}     & {3.0}    & {5.0}    &  & {1.0}    & {2.0}    & {3.0}   &                       &                       &                       &                              \\ \hline
Identity                    & {99.7}                         & {38.2}    & {84.2}   & {93.1}   &  & {17.9}    & {94.3}   & {96.2}   &  & {29.6}   & {75.6}   & {90.3}  & 99.3                  & 98.9                  & 6.50                  & 6.31                         \\
Gender                      & {99.8}                         & {73.2}    & {87.0}   & {95.7}   &  & {65.9}    & {96.2}   & {98.6}   &  & {63.0}   & {78.4}   & {88.5}  & 99.6                  & 96.3                  & 62.0                  & 62.0                         \\ \Xhline{2\arrayrulewidth}
\end{tabular}%
}
        \caption*{(a) MPIIFaceGaze}
    \end{subtable}
    \vspace{-0.15in}
    \begin{subtable}{}
        \centering
        \resizebox{\textwidth}{!}{%
\begin{tabular}{ccccccccccccccccc}
\Xhline{2\arrayrulewidth}
\multirow{2}{*}{Attributes} & \multirow{2}{*}{{w/o Defense}} & \multicolumn{3}{c}{{GauDP}} &  & \multicolumn{3}{c}{{IP-DP}} &  & \multicolumn{3}{c}{{FS-DP}} & \multirow{2}{*}{MaxP} & \multirow{2}{*}{BDAP} & \multirow{2}{*}{TPGD} & \multirow{2}{*}{PrivateGaze} \\ \cline{3-5} \cline{7-9} \cline{11-13}
                            &                              & {0.1}     & {0.3}    & {0.5}    &  & {0.3}     & {3.0}    & {5.0}    &  & {1.0}    & {2.0}    & {3.0}   &                       &                       &                       &                              \\ \hline
Identity                    & {99.9}                         & {4.85}    & {95.1}   & {99.6}   &  & {6.20}    & {86.4}   & {97.4}   &  & {5.86}   & {38.9}   & {73.8}  & 99.7                  & 91.1                  & 2.82                  & 1.13                         \\
Gender                      & {99.9}                         & {57.5}    & {87.2}   & {96.9}   &  & {57.0}    & {96.9}   & {99.2}   &  & {57.6}   & {67.7}   & {82.2}  & 99.5                  & 95.6                  & 56.9                  & 56.9                         \\ \Xhline{2\arrayrulewidth}
\end{tabular}%
}
        \caption*{(b) ColumbiaGaze}
    \end{subtable}
    \vspace{-0.1in}
\end{table}

For GauDP, IP-DP, and FS-DP, we observe that the identity and gender recognition accuracies decrease as the values of $\epsilon$ drop, indicating that these methods can effectively preserve user privacy with smaller $\epsilon$ values. For example, the identity recognition accuracy for GauDP with $\epsilon=0.1$ is $4.85\%$ on the ColumbiaGaze dataset, while for FS-DP is $5.86\%$ with $\epsilon=1.0$.
However, as we will demonstrate in Section \ref{subsec:utilityPerformance}, the utility performance of GauDP, IP-DP, and FS-DP with small $\epsilon$ values is significantly worse compared to \name.

For obfuscated images obtained by MaxP and B-DAP, the identity recognition accuracy exceeds 90\%, and the gender recognition accuracy is higher than 95\% on both datasets. This outcome implies that the obfuscated images generated by MaxP and B-DAP still retain discernible features, which could potentially be exploited by attackers to infer the private attributes of the users.
Lastly, TPGD achieves results similar to \name on both testing datasets, as it is designed based on our framework.

\begin{figure}[]
  \centering
  \includegraphics[width=0.79\linewidth]{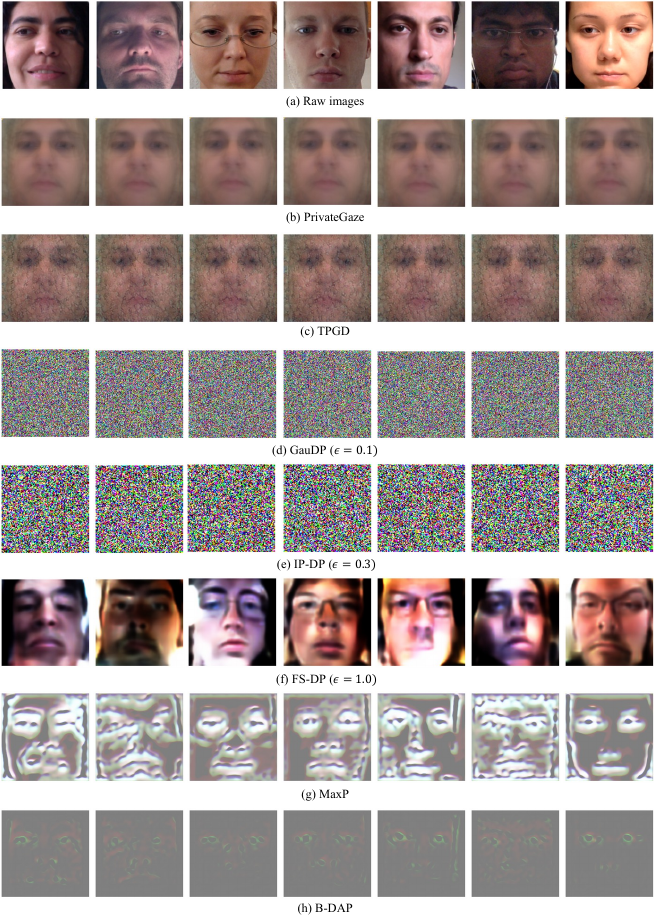}
  \caption{Illustration of (a) raw images of different subjects and obfuscated images generated by (b) \name, (c) TPGD, {(d) GauDP ($\epsilon=0.1$), (e) IP-DP ($\epsilon=0.3$), (f) FS-DP ($\epsilon=1.0$)}, (g) MaxP, and (h) B-DAP. The obfuscated images obtained by \name and TPGD have similar appearances, making it challenging for attackers to infer user identity and gender from the obfuscated images.}
  \label{Fig:ConvertedImages}
  \vspace{-0.1in}
\end{figure}

\subsubsection{In-depth analysis and discussion.}
\label{subsubsec:privacydiscussion}
As shown in Table \ref{Tab:privacy}, the gender recognizers for \name and TPGD have the same recognition accuracy, i.e., 62.0\% and 56.9\% on MPIIFaceGaze and ColumbiaGaze datasets, respectively. We observe that when obfuscated images contain no discernible gender cues, the trained recognizers tend to output the same gender label for any inputs. Essentially, without gender-related information in the obfuscated images, the best knowledge the recognizers can obtain is an approximation of the gender distribution in the training dataset $\mathcal{D}_p$. Consequently, during inference, leveraging prior probability distributions, the recognizers predict the gender of a given testing image to be the one with the highest prior probability, thereby achieving a better classification accuracy than random guessing.

For instance, in the MPIIFaceGaze dataset, the gender recognizers for both \name and TPGD consistently classify subjects in all testing images as male. This outcome results from the imbalanced gender distribution within the training dataset $\mathcal{D}_p$, where 59.5\% of images feature male subjects. Consequently, given that 62.0\% of the testing images contain male subjects, this results in an equal gender recognition accuracy of 62.0\%. Thus, these evaluation results demonstrate the effectiveness of \name in preserving gender information.

In contrast to the imbalanced gender distribution, the identity distribution within the training set $\mathcal{D}_p$ for both MPIIFaceGaze and ColumbiaGaze datasets is well-balanced, with each subject having an equal number of training images. Thus, when the identity recognizers cannot learn any identity-related information from the obfuscated images and are faced with equal prior probabilities, the identity recognition accuracy is similar to that of random guess, i.e., 6.31\% and 1.13\% for MPIIFaceGaze and ColumbiaGaze datasets, respectively. This result further demonstrates the capability of \name in preserving user information from potential malicious service providers.

\subsubsection{Visualization of obfuscated images.} 

We also perform a visual comparison between the raw full-face images and the obfuscated images generated by the proposed \name and other baseline methods. As shown in Figures~\ref{Fig:ConvertedImages} (b) and (c), the obfuscated images generated by \name and TPGD exhibit similar visual characteristics among themselves, making it extremely challenging for an attacker to infer the user private attributes, even when correctly labeled obfuscated images are used to train the deep learning-based classifiers (as showcased in Table~\ref{Tab:privacy}). In contrast, as shown in Figure~\ref{Fig:ConvertedImages} (g), the user attributes are much more discernible in the obfuscated images produced by MaxP. Moreover, despite it might be difficult for human observers to identify the subjects in images obfuscated by B-DAP, the results shown in Table \ref{Tab:privacy} indicate that an attacker can successfully train a classifier on these obfuscated images to accurately classify user private attributes.

As shown in Figures~\ref{Fig:ConvertedImages} (d) and (e), GauDP ($\epsilon=0.1$) and IP-DP ($\epsilon=0.3$) apply significant perturbations to the raw images, making it difficult for the malicious service provider to classify identities and genders from them. %obfuscated images. 
Furthermore, FS-DP ($\epsilon=1.0$) generates obfuscated images that have significantly different appearances from raw images by perturbing features with strong random noises, effectively preserving user privacy.

\subsubsection{Residual map.} 
\begin{figure}[]
\centering
  \includegraphics[width=0.76\linewidth]{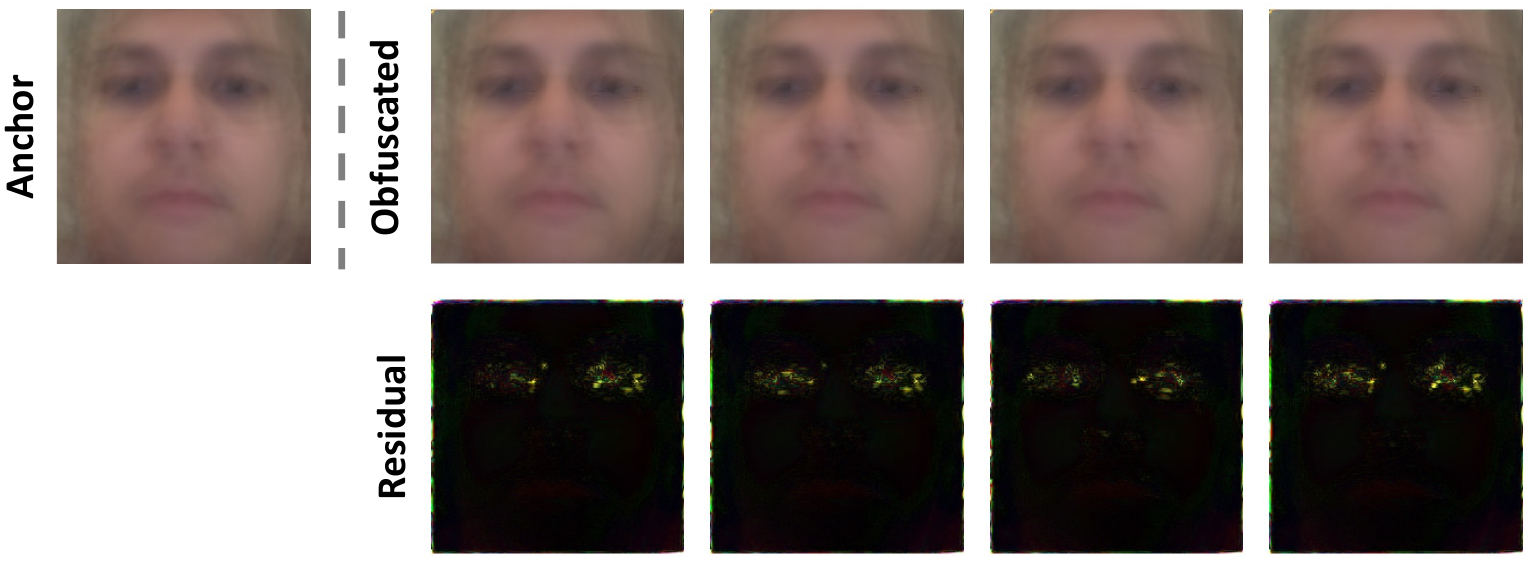}
  \caption{Illustration of the residual maps (scaled by a factor of 30) generated from the obfuscated images (with different gaze directions) compared to the anchor image. The privacy preserver introduces perturbations into the eye regions of the anchor image when generating obfuscated images, resulting in different gaze directions.}
\label{Fig:residualMap}
% \vspace{-0.1in}
% \end{minipage}
\end{figure}

To better understand why the obfuscated images generated by \name have similar appearances yet lead the black-box gaze estimators to output varied gaze directions, we examine the residual maps between the obfuscated image and the anchor image. As shown in Figure~\ref{Fig:residualMap}, the privacy preserver adds perturbations to the eye regions of the anchor image, which are the most critical parts of the facial image for gaze estimation~\cite{zhang17_cvprw}. These perturbations are learned by the privacy preserver and contain the gaze-related features of the raw images. In this way, the obfuscated images retain similar appearances while leading to different gaze directions when fed into the black-box gaze estimator.

\subsection{Performance in the Utility Goal}
\label{subsec:utilityPerformance}

Below, we evaluate the performance of different methods in achieving the utility goal, i.e., gaze estimation task. The results are reported in Table \ref{Tab:utility}, which demonstrate that \name consistently outperforms all compared methods and achieves the lowest average angular error across two datasets and with five different neural network architectures for the black-box gaze estimator. On average, \name improves gaze estimation performance by 41.79\%-78.47\%, and 34.13\%-64.68\% on MPIIFaceGaze and ColumbiaGaze, respectively.

\begin{table}
    \centering
    \caption{Performance in the utility goal is measured by the average angular error (in degree) of the black-box gaze estimator across five different structures, using obfuscated images generated by different methods as inputs for gaze estimation. {We present results for GauDP, IP-DP, and FS-DP with different $\epsilon$ values, respectively.} \name consistently outperforms all compared methods in all examined scenarios. Overall, \name achieves an average performance improvement in gaze estimation of 49.86\%-77.10\%, and 34.13\%-60.67\% on MPIIFaceGaze and ColumbiaGaze, respectively.}
    \vspace{-0.15in}
    \label{Tab:utility}
    \begin{subtable}{}
        \centering
        \resizebox{\textwidth}{!}{%
\begin{tabular}{cccccccccccccccc}
\Xhline{2\arrayrulewidth}
\multirow{2}{*}{Sructures} & \multicolumn{3}{c}{{GauDP}} &  & \multicolumn{3}{c}{{IP-DP}} &  & \multicolumn{3}{c}{{FS-DP}} & \multirow{2}{*}{MaxP} & \multirow{2}{*}{B-DAP} & \multirow{2}{*}{TPGD} & \multirow{2}{*}{PrivateGaze} \\ \cline{2-4} \cline{6-8} \cline{10-12}
                           & {0.1}     & {0.3}   & {0.5}     &  & {0.3}     & {3.0}    & {5.0}    &  & {1.0}     & {2.0}    & {3.0}    &                       &                        &                       &                              \\ \hline
ResNet18                   & {20.73}   & {20.37}  & {20.09}  &  & {20.94}   & {26.22}  & {28.90}  &  & {16.09}   & {15.63}  & {14.90}  & 13.65                 & 20.07                  & 17.28                 & \textbf{7.02}                         \\
MobileNetV2                & {25.70}   & {25.72} & {25.69}   &  & {30.03}   & {30.29}  & {30.15}  &  & {19.86}   & {18.16}  & {16.46}  & 29.95                 & 46.52                  & 17.82                 & \textbf{6.66}                         \\
ShuffleNet                 & {13.93}   & {13.62} & {13.33}   &  & {13.77}   & {15.22}  & {15.50}  &  & {19.68}   & {17.95}  & {16.16}  & 18.93                 & 23.38                  & 11.71                 & \textbf{8.34}                         \\
VGG11                      & {14.06}   & {14.82} & {15.77}   &  & {14.32}   & {15.76}  & {16.64}  &  & {20.99}   & {19.63}  & {18.23}  & 17.72                 & 13.43                  & 15.89                 & \textbf{7.50}                         \\
EfficientNet               & {86.02}   & {44.52} & {26.51}   &  & {91.68}   & {69.92}  & {40.25}  &  & {19.36}   & {17.36}  & {15.58}  & 13.32                 & 13.08                  & 9.99                  & \textbf{7.23}                         \\ \hline\\[-2ex]  \hline \\[-2ex]
Average                    & {32.09}   & {23.81} & {20.28}   &  & {34.14}   & {31.48}  & {26.28}  &  & {19.19}   & {17.74}  & {16.26}  & 18.71                 & 23.29                  & 14.66                  & \textbf{7.35}                         \\
Improvement                & {77.10\%} & {69.13\%}& {63.75\%}&  & {78.47\%} & {76.65\%}& {72.03\%} &  & {41.79\%} & {58.56\%}& {54.79\%}& 60.72\%               & 68.44\%                & 49.86\%                &                              \\ \Xhline{2\arrayrulewidth}
\end{tabular}%
}
        \caption*{(a) MPIIFaceGaze}
    \end{subtable}
    \vspace{-0.15in}
    \begin{subtable}{}
        \centering
        \resizebox{\textwidth}{!}{%
\begin{tabular}{cccccccccccccccc}
\Xhline{2\arrayrulewidth}
\multirow{2}{*}{Sructures} & \multicolumn{3}{c}{{GauDP}} &  & \multicolumn{3}{c}{{IP-DP}} &  & \multicolumn{3}{c}{{FS-DP}} & \multirow{2}{*}{MaxP} & \multirow{2}{*}{B-DAP} & \multirow{2}{*}{TPGD} & \multirow{2}{*}{PrivateGaze} \\ \cline{2-4} \cline{6-8} \cline{10-12}
                           & {0.1}     & {0.3}   & {0.5}     &  & {0.3}     & {3.0}    & {5.0}    &  & {1.0}     & {2.0}    & {3.0}    &                       &                        &                       &                              \\ \hline
ResNet18                   & {14.46}   & {14.89} & {15.65}   &  & {19.42}   & {26.75}  & {30.80}  &  & {19.35}   & {18.90}  & {18.80}  & 15.42                 & 15.22                  & 21.45                 & \textbf{9.72}                         \\
MobileNetV2                & {17.14}   & {17.27} & {17.33}   &  & {20.55}   & {21.02}  & {21.03}  &  & {19.30}   & {18.96}  & {18.71}  & 22.69                 & 41.32                  & 23.17                 & \textbf{11.45}                        \\
ShuffleNet                 & {13.58}   & {13.48} & {13.31}   &  & {12.86}   & {12.55}  & {12.36}  &  & {19.94}   & {20.44}  & {20.26} & 20.96                 & 24.87                  & 19.35                 & \textbf{11.19}                        \\
VGG11                      & {14.60}   & {13.75}  & {13.39}   &  & {18.42}   & {15.78}  & {15.11} &  & {24.62}   & {23.54}  & {22.99} & 27.24                 & 18.50                  & 20.81                 & \textbf{12.19}                        \\
EfficientNet               & {83.52}   & {44.61} & {25.86}   &  & {86.91}   & {63.39}  & {36.51}  &  & {21.95}   & {21.39}  & {20.68} & 15.72                 & 12.64                  & 13.67                 & \textbf{11.80}                        \\ \hline\\[-2ex]  \hline \\[-2ex]
Average                    & {28.66}   & {20.8}  & {17.11}   &  & {31.63}   & {27.89}  & {23.16}  &  & {21.03}   & {20.64}  & {20.29} & 20.40                 & 22.50                  & 19.69                 & \textbf{11.17}                        \\
Improvement                & {60.67\%} & {43.75\%}& {34.13\%}&  & {64.68\%} & {59.94\%}& {51.77\%}&  & {46.88\%} & {45.88\%}& {44.94\%}& 44.75\%               & 49.91\%                & 42.76\%             &                              \\ \Xhline{2\arrayrulewidth}
\end{tabular}%
}
        \caption*{(b) ColumbiaGaze}
    \end{subtable}
\vspace{-0.1in}    
\end{table}

For differential privacy-based methods, i.e., GauDP, IP-DP, and FS-DP, while they achieve strong privacy performance by setting $\epsilon$ to small values, their utility performance notably lags behind \name. For instance, GauDP with $\epsilon=0.1$ results in an average angular error over various structures of $\mathcal{G}_b(\cdot)$ of 32.09$^\circ$ on MPIIFaceGaze, nearly five times higher than that of \name. Increasing the values of $\epsilon$ improves the utility performance of differential privacy-based methods by reducing the intensity of the added perturbation for privacy protection. However, as shown in Table~\ref{Tab:privacy}, this improvement comes at the cost of deteriorating privacy performance.

On MPIIFaceGaze dataset, the average utility performance for MaxP and B-DAP is 18.71$^\circ$ and 23.29$^\circ$ respectively. These results indicate that the autoencoder and image converter trained by MaxP and B-DAP, respectively, with the surrogate gaze estimator cannot generalize well to black-box gaze estimators. Despite TPGD achieving similar performance in privacy protection as \name (as shown previously in Table~\ref{Tab:privacy}), its average utility performance over different black-box gaze estimators is 14.66$^\circ$, which is significantly worse than that for \name, at 7.35$^\circ$. 
Similar observations hold for the evaluation results on the ColumbiaGaze dataset.

\begin{figure}[t]
	\centering
	\subfigure[MPIIFaceGaze]{\includegraphics[scale=0.25]{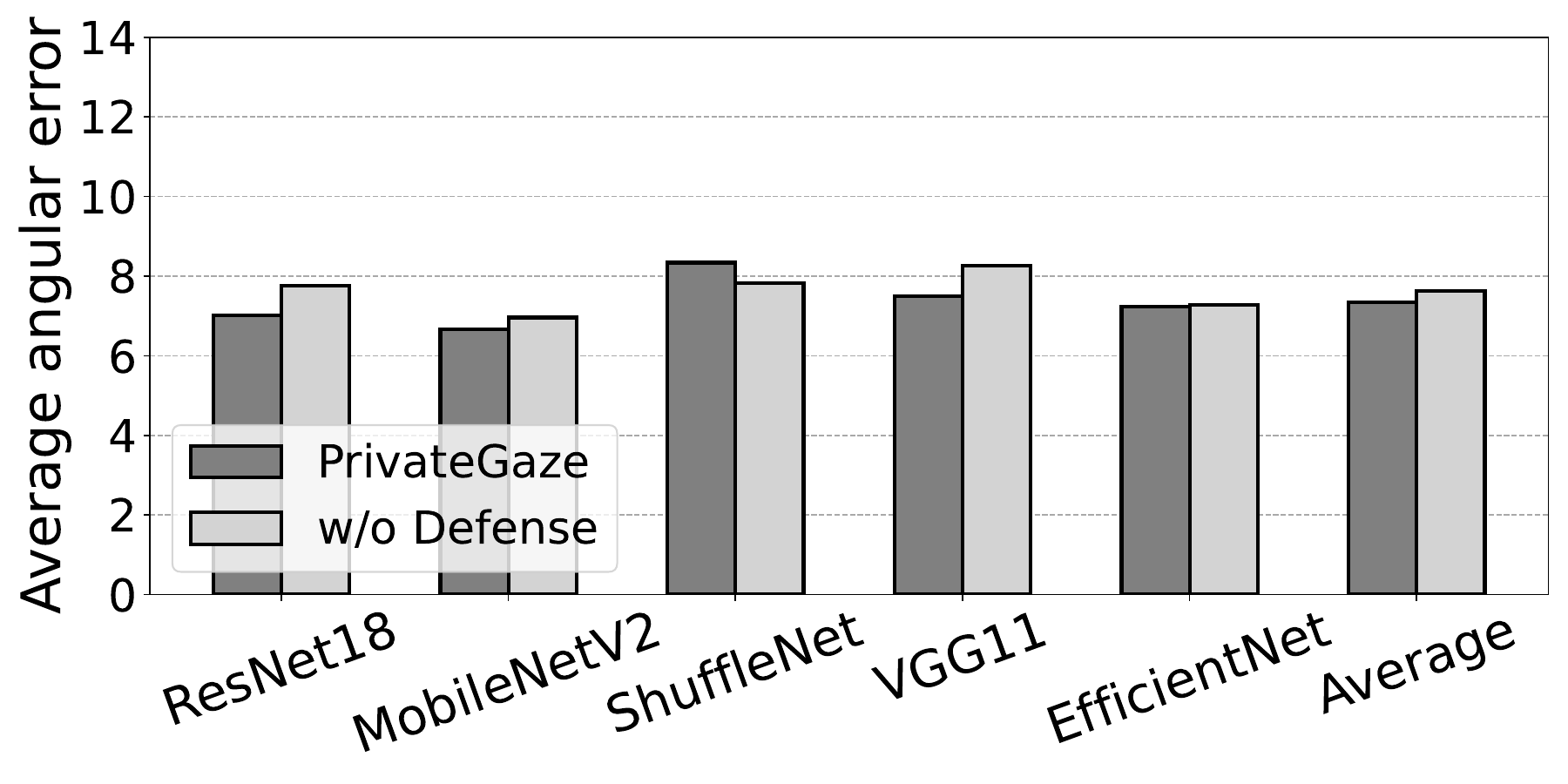}\label{Fig:PreGaze}} 
 \subfigure[ColumbiaGaze]{\includegraphics[scale=0.25]{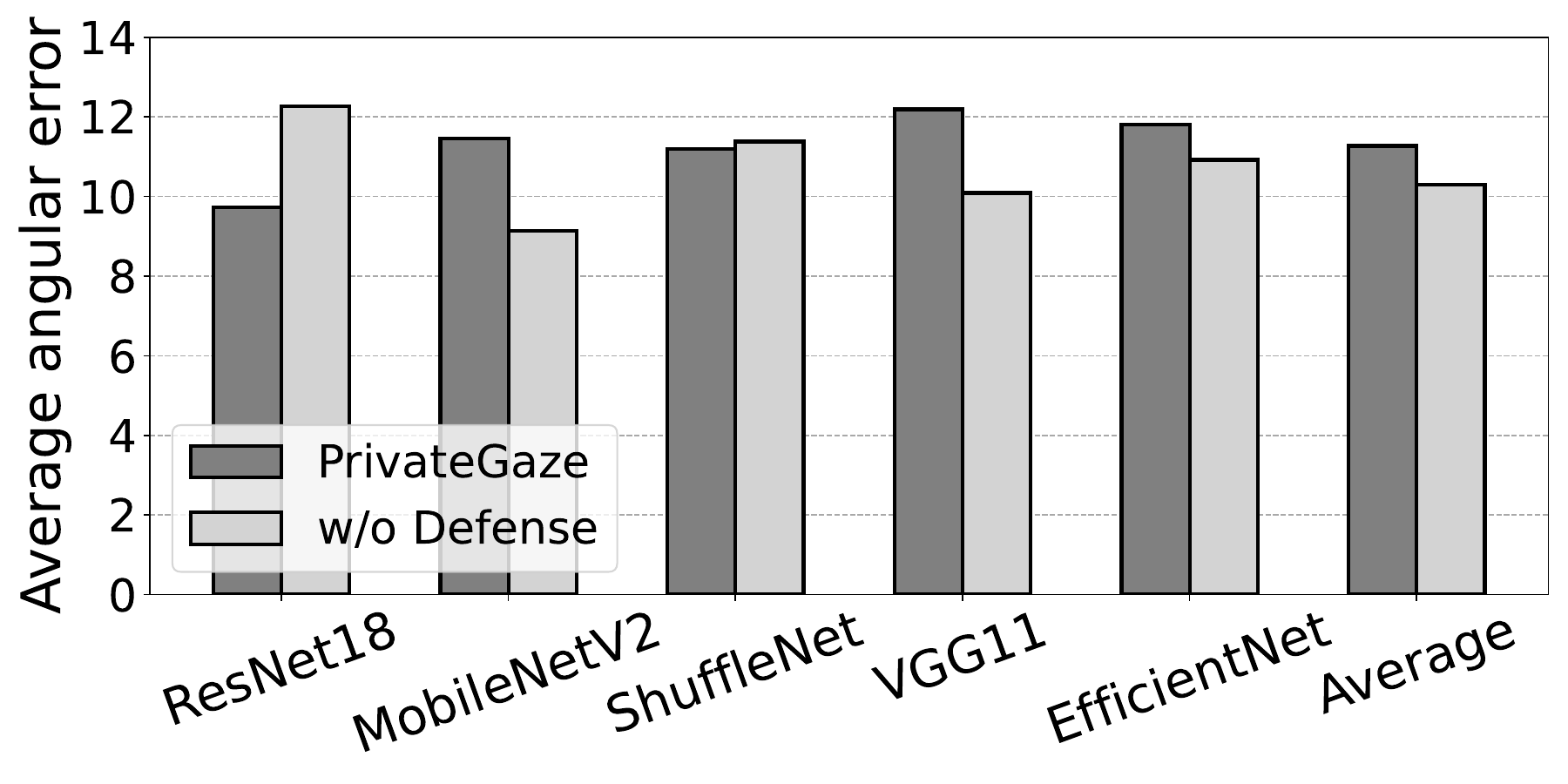}\label{Fig:TPGD}}
	
\caption{Average angular error (in degree) of the black-box gaze estimator when using obfuscated images generated by \name compared to raw images, i.e., w/o Defense, as inputs for gaze estimation. Our method demonstrates performance comparable to that of raw images across different structures and datasets. %Note that in all cases, the black-box gaze estimator is trained on raw images without any perturbations.
}
\label{Fig:UtilityComparsion}
\vspace{-0.1in}
\end{figure}

To further investigate the performance of \name on the utility goal, we compare the average angular error of the black-box gaze estimator when using obfuscated images generated by \name versus raw images (\emph{w/o Defense}) as the inputs.
The results are shown in Figure~\ref{Fig:UtilityComparsion}. On the MPIIFaceGaze dataset, the average performance of \name among different structures is 7.35$^\circ$, which is better than that of \textit{w/o Defense} at 7.62$^\circ$. On the ColumbiaGaze dataset, \name achieves an average utility performance of 11.27$^\circ$ across different structures, which is slightly higher than \textit{w/o Defense} at 10.29$^\circ$. Overall, these results indicate that \name maintains comparable utility performance for the black-box gaze estimator even when compared to a method that does not employ any privacy protection.

\subsubsection{Discussion on utility performance}
\label{subsubsec:discussion_utility}
As shown in Figure \ref{Fig:UtilityComparsion}, on the MPIIFaceGaze dataset, \name achieves superior gaze estimation performance compared to \emph{w/o Defense} on average. To explore the reason behind this improvement, we observed a reduction in the average angular error of $\mathcal{G}_w(\cdot)$ during training on obfuscated images generated by $\mathcal{P}(\cdot)$ from the validation set. We observe a decrease from 12$^\circ$ to 3.52$^\circ$, which is lower than the error observed on raw images, at 5.31$^\circ$. This suggests that the privacy preserver acts as an \textit{image filter} that eliminates redundant features from raw images, thereby enhancing gaze estimation performance.

Moreover, we observe that the average gaze estimation performance of \name is better than \emph{w/o Defense} on MPIIFaceGaze, while it shows marginally inferior performance on ColumbiaGaze. This difference can be attributed to the training data used for the privacy preserver, which was trained on the GazeCapture dataset and evaluated on both MPIIFaceGaze and ColumbiaGaze. As shown in Figure~\ref{Fig:DatasetIllustration}, both GazeCapture and MPIIFaceGaze are acquired under real-world conditions employing front-facing cameras in mobile devices, whereas ColumbiaGaze is collected in a more controlled laboratory environment using web cameras. The images from GazeCapture are more similar to those from MPIIFaceGaze than to those from ColumbiaGaze. Consequently, the image filter, i.e., privacy preserver, operates more effectively on MPIIFaceGaze compared to ColumbiaGaze, and leads to better gaze estimation performance on MPIIFaceGaze.

\subsubsection{In-depth analysis on the generalization ability of \name.}
\label{subsubsec:discussion_generalization}
\name demonstrates good generalization ability, i.e., $\mathcal{P}(\cdot)$ is trained with $\mathcal{G}_w(\cdot)$ but yields good gaze estimation performance on $\mathcal{G}_b(\cdot)$. We attribute this capability primarily to the proposed anchor image generation module and privacy protection mechanism. Specifically, the privacy protection mechanism ensures that the obfuscated images $x'$ closely resemble the anchor image $\hat{x}$, resulting in similar gaze estimation results for both $\mathcal{G}_w(\cdot)$ and $\mathcal{G}_b(\cdot)$. This alignment encourages the obfuscated images to produce similar outputs for both $\mathcal{G}_w(\cdot)$ and $\mathcal{G}_b(\cdot)$. Consequently, while \name optimizes the $x'$ to facilitate accurate gaze direction inference by $\mathcal{G}_w(\cdot)$, it also achieves good gaze estimation performance on $\mathcal{G}_b(\cdot)$. Our experiments validate this analysis. In contrast, TPGD utilizes the anchor image to generate obfuscated images without considering their similarity to the anchor image, leading to substantially inferior performance compared to \name in terms of utility. Moreover, as detailed in Section \ref{subsubsec:ablationStudyAnchor}, neglecting the outputs of $\mathcal{G}_w(\cdot)$ and $\mathcal{G}_b(\cdot)$ during the generation of $\hat{x}$ results in a significant performance decline for \name in terms of utility.

\subsection{Overall Performance Comparison}
\label{subsec:overallPerformance}

\begin{figure}[]
	\centering
	\subfigure[MPIIFaceGaze]{\includegraphics[scale=0.46]{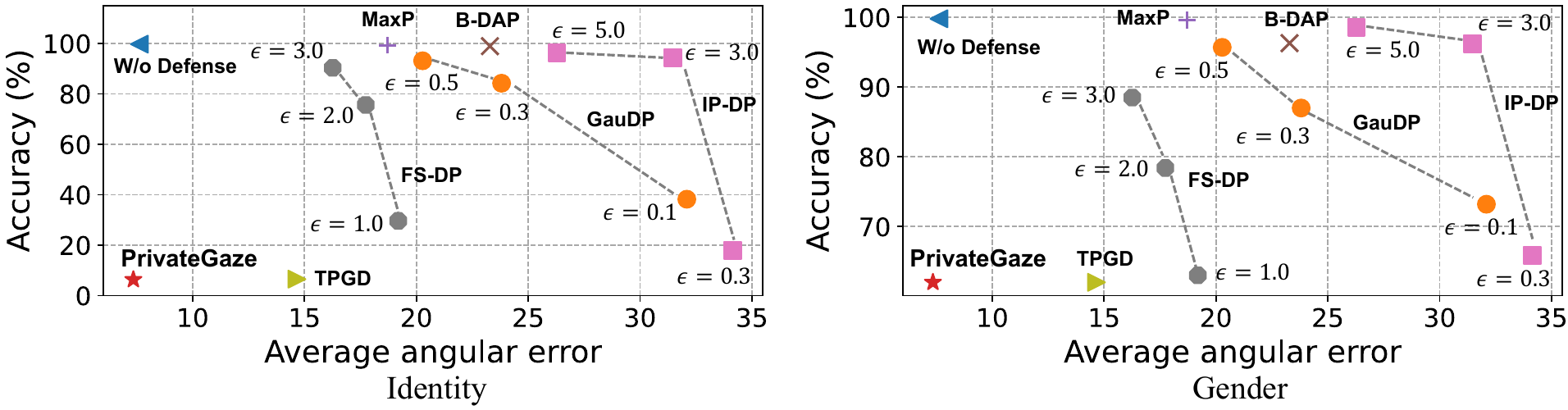}\label{Fig:OverallPerforMPII}} 
 \subfigure[ColumbiaGaze]{\includegraphics[scale=0.46]{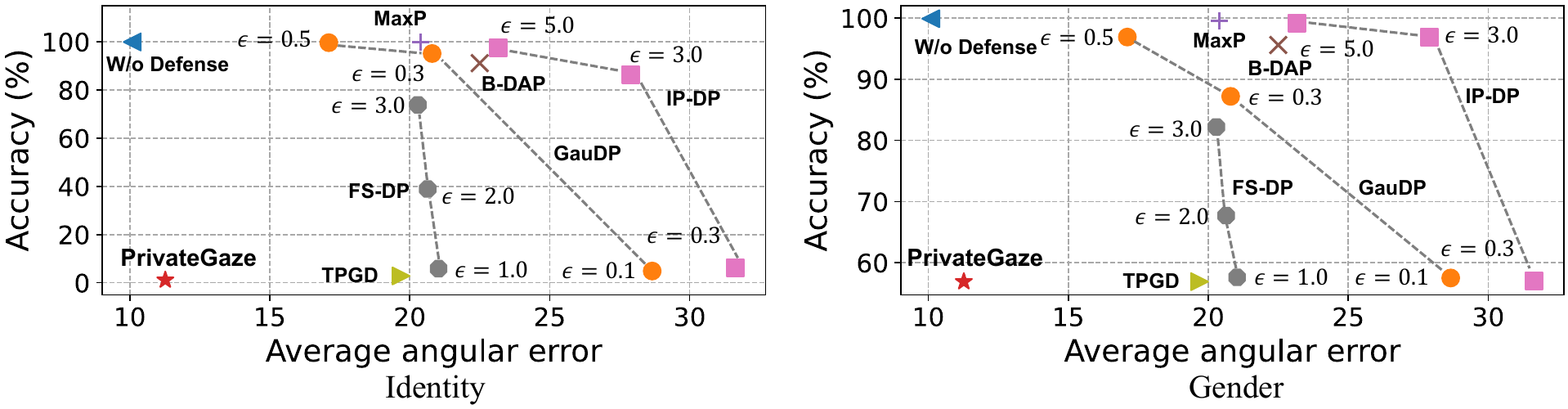}\label{Fig:OverallPerforColu}} 

\caption{{Overall performance comparison between \name and the compared methods on (a) MPIIFaceGaze and (b) ColumbiaGaze datasets. The X-axis is the utility performance, i.e., the average angular error, while the Y-axis is the privacy performance, i.e., the identity recognition accuracy and gender recognition accuracy. The plots are the identity (gender) recognition accuracy and the average angular error for \name and the compared methods. The overall performance of \name lies on the lower left corner in all the evaluation scenarios, which indicates the superiority of \name on the privacy-utility trade-off.}
}
\label{Fig:OverallPerformance}
\vspace{-0.1in}
\end{figure}

Below, we compare the overall performance, i.e., the privacy-utility trade-off, between \name and the compared methods. The results are shown in Figure \ref{Fig:OverallPerformance}. Across all evaluation scenarios, \name consistently occupies the lower left corner, indicating its effectiveness in preserving user privacy while maintaining good utility performance. Notably, TPGD, GauDP ($\epsilon=0.1$), FS-DP ($\epsilon=1.0$), and IP-DP ($\epsilon=0.3$) demonstrate privacy-preserving performance comparable to \name, yet their utility performances significantly lag behind. Moreover, among the differential privacy-based methods, i.e., GauDP, IP-DP, and FS-DP, FS-DP achieves the most favorable privacy-utility trade-off. This is because FS-DP adopts state-of-the-art DP techniques and involves the gaze estimator in the training stage. Lastly, MaxP and B-DAP exhibit overall performances within the upper middle region, indicating their inability to achieve both utility and privacy goals simultaneously.

\subsection{Ablation Studies}

We conduct ablation studies to investigate the impact of different design choices on the system performance. We use MPIIFaceGaze as the testing set and implement a black-box gaze estimator using EfficientNet.

\subsubsection{Impact of $\lambda$ on the performance trade-off between utility and privacy}
\label{subsubsec:ablationLambda}
\begin{figure}[t]
	\centering
	\subfigure[Utility]{\includegraphics[scale=0.25]{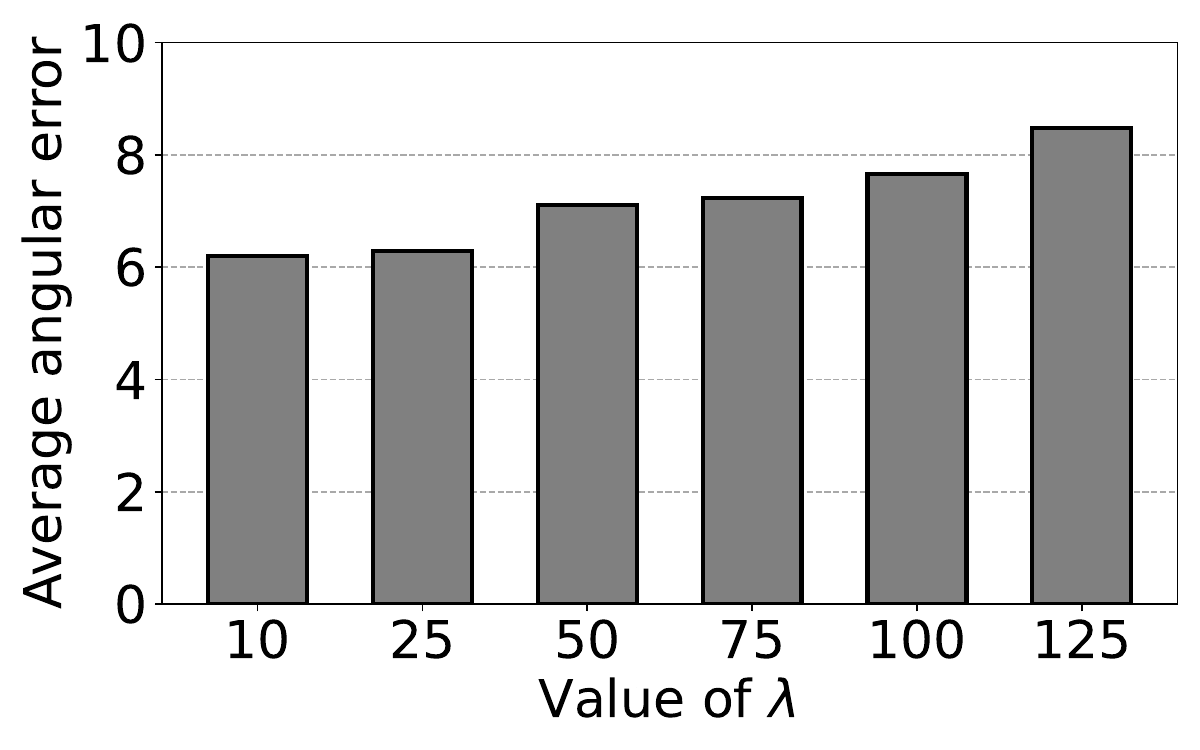}\label{Fig:ablationUtility}} 
 \subfigure[Preserving users identities]{\includegraphics[scale=0.25]{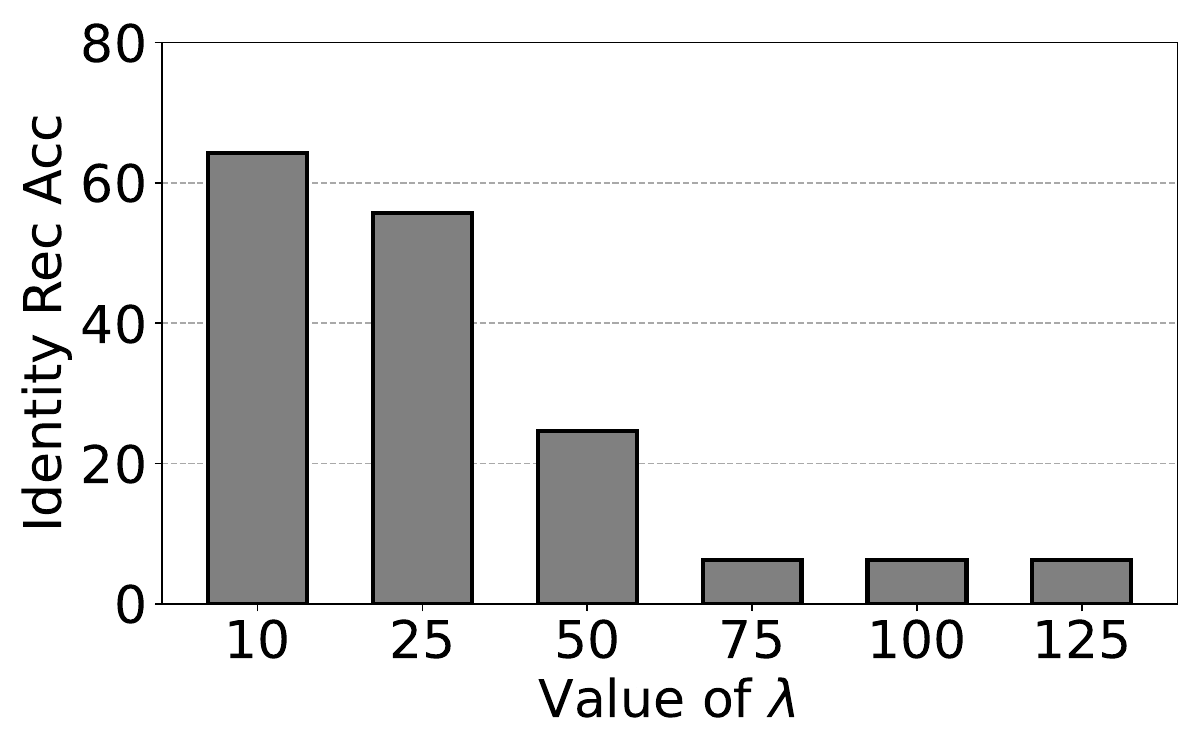}\label{Fig:ablationID}}
 \subfigure[Preserving users genders]{\includegraphics[scale=0.25]{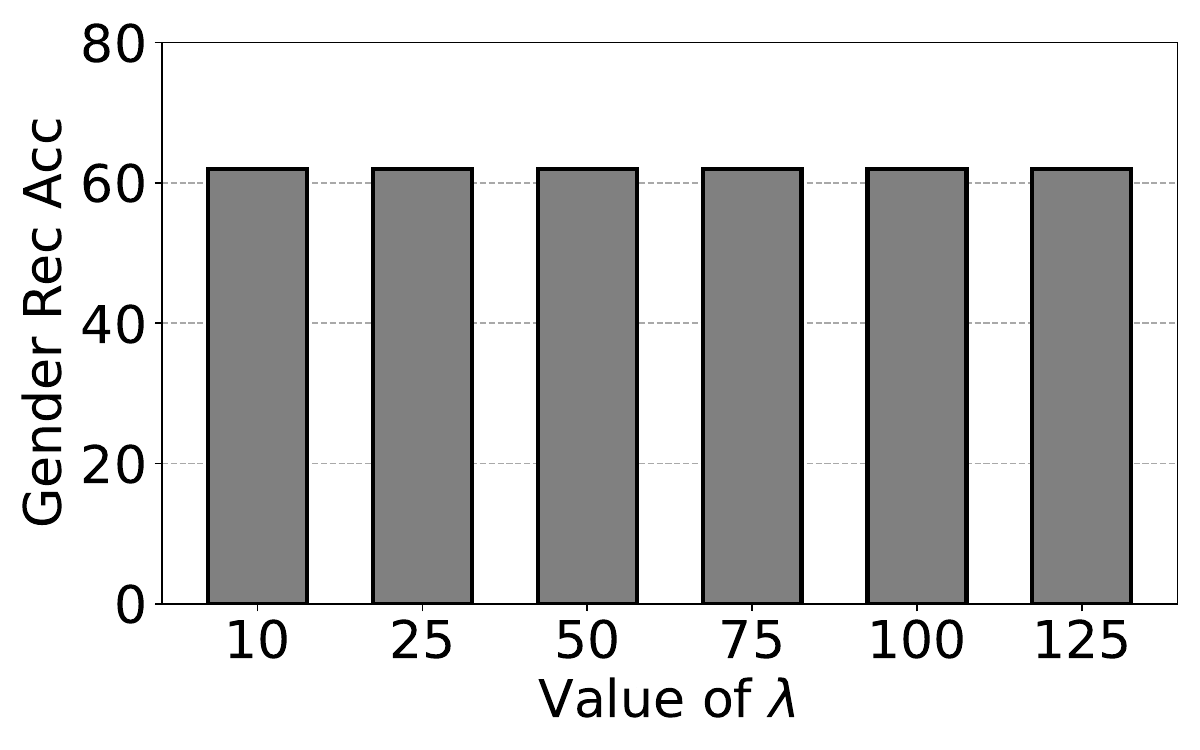}\label{Fig:ablationGender}}

\caption{Impact of $\lambda$ on the performance trade-off between utility and privacy. (a) utility, (b) preserving users' identities, and (c) preserving users' genders. The value of $\lambda$ trades off the performance of \name on utility against privacy. \name sets $\lambda=75$, which allows \name to consistently preserve user privacy while maintaining comparable utility performance to raw images across different datasets.}
	\label{Fig:AblationLambda}
\end{figure}

For the optimization problem described in Equation \ref{Eq:loss_function}, the parameter $\lambda$ trades off the utility objective and the privacy-preserving objective. To explore how changes in $\lambda$ affect the performance of \name, we experiment with varying $\lambda$ within the range from $10$ to $125$ and present results in Figure~\ref{Fig:AblationLambda}.

Overall, reducing the value of $\lambda$ improves the utility performance of \name yet degrades its privacy performance when $\lambda$ is set below 75. Particularly, there is a decrease in the average angular error from 8.48$^\circ$ to 6.10$^\circ$ when $\lambda$ is reduced from 125 to 10. This indicates an increased weighting of the utility objective in the optimization problem. For privacy objective, the identity recognition accuracy is higher than 20\% when the value of $\lambda$ is smaller than 75, while the gender recognition accuracy is stable for all the examined values of $\lambda$. In conclusion, we have chosen to set $\lambda$ to 75 in our implementation. This setting effectively balances the need to preserve user privacy in the obfuscated images while maintaining comparable utility performance to raw images across various datasets.

\begin{table}[]
\centering
%\captionsetup{font={small,stretch=1.}, justification=raggedright}

\caption{Impact of the proposed anchor image generation algorithm on the performance of \name. Our method of generating anchor images is essential for maintaining the utility performance of \name.}
\resizebox{0.67\textwidth}{!}{%
\begin{tabular}{ccc}
\Xhline{2\arrayrulewidth}
Task           & {\name}      & RandomAnchor  \\ \hline
Utility: Gaze estimation          &  7.23$^\circ$        & 9.06$^\circ$         \\
Privacy: Identity recognition accuracy &  6.31\%        & 6.40\%       \\
Privacy: Gender recognition accuracy   &  62.0\%        & 62.0\%        \\ \Xhline{2\arrayrulewidth}
\end{tabular}
}
\vspace{-0.1in}
\label{Tab:albationAnchorImage}
\end{table}

\subsubsection{Impact of anchor image generation}
\label{subsubsec:ablationStudyAnchor}

We investigate the impact of the proposed anchor image generation module (described in Section~\ref{subsubsec:anchorImageGeneration} and Algorithm~\ref{Alg:anchorImage}) on the performance of \name. Specifically, instead of querying the black-box gaze estimator and the surrogate gaze estimator to find a suitable set of images, we randomly sample 50 images from the training set to form an average facial image as the anchor image. We use the term \textit{RandomAnchor} to denote the method of generating anchor images through %the use of
randomly sampled images. The results are shown in Table \ref{Tab:albationAnchorImage}. 

First, the anchor image does not affect the performance of \name on the privacy objective, as the identity recognition accuracy and the gender recognition accuracy of RandomAnchor and \name are similar. Second, the average angular error of RandomAnchor is 20\% higher than that of \name. This indicates that the proposed method for generating the anchor image can improve the generalizability of the privacy preserver. In other words, when trained with the surrogate gaze estimator, the privacy preserver can still achieve good gaze estimation performance when the obfuscated images are used by the black-box gaze estimator. 
Moreover, we observe that the utility performance of RandomAnchor is superior to that of the other baseline methods shown in Table \ref{Tab:utility}. This highlights the effectiveness of our method in achieving the utility objective. 

\subsubsection{Impact of gender of the anchor image on preserving gender information}

In general, the anchor image is constructed by averaging full-face images from both male and female subjects, ensuring it does not portray a specific gender. When using the anchor image as the base in generating the obfuscated images, as illustrated in Figure \ref{Fig:residualMap}, the resulting obfuscated images will resemble the anchor image and also avoid portraying a specific gender. Nevertheless, there is a possibility that the images used to generate the anchor image predominantly belong to subjects of a specific gender, especially in cases where the gender distribution within the training dataset is imbalanced. In such instances, the anchor image may inadvertently exhibit characteristics of that specific gender.

\begin{figure}[]
    \centering
    \subfigure[Female]{\includegraphics[scale=0.3]{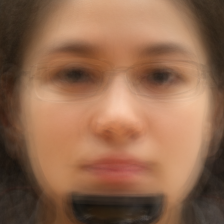}}
    \hspace{50pt}
    \subfigure[Male]{\includegraphics[scale=0.3]{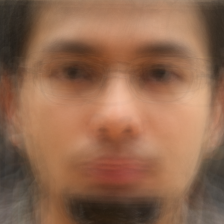}} 
    \caption{{Two anchor images that are generated using images containing (a) female-only subjects and (b) male-only subjects, respectively.}}
    \label{fig:male_female_anchor}
    \vspace{-0.15in}
\end{figure}

To study how the gender of the anchor image affects the performance of \name in preserving gender information, we generate an anchor image $\hat{x}_f$ {using only images containing female subjects}. The resulting anchor image is shown in Figure \ref{fig:male_female_anchor} (a). In this case, we consider the genders of $\hat{x}_f$ and the corresponding obfuscated images as ``female''. 
We find that the gender classification accuracy on the obfuscated images remains at 62.0\%. As discussed in Section 4.5.1, this indicates the effectiveness of \name in preserving user's gender information. Similarly, we conduct experiments using only male images to generate the anchor image (illustrated in Figure \ref{fig:male_female_anchor} (b)), and we obtain the same result. These results demonstrate that the efficacy of \name in preserving users' gender information is not compromised even when the anchor image exhibits a specific gender.

\subsection{System Performance on Different Computation Platforms}
% To demonstrate the system performance of the proposed \name, 
We measure the processing time and memory usage of the privacy preserver when implemented with different architectures and deployed on different computation platforms. In addition to the default structure, which consists of four convolutional and up-convolutional blocks, we also evaluate a variant with such blocks. We assess the performance of \name on three hardware platforms, including a desktop equipped with an NVIDIA GeForce RTX 3080Ti GPU, a laptop featuring an NVIDIA GeForce RTX 3060 GPU, and a laptop equipped with an NVIDIA GeForce RTX 1050Ti GPU. These platforms are chosen to represent a wide range of common computational devices used in daily scenarios.

\vspace{0.05in}
\noindent\textbf{Processing time.} 
We measure the latency introduced by the privacy preserver in generating obfuscated images. We randomly sample one image from the MPIIFaceGaze dataset and feed it into the privacy preserver. We repeat the experiment 1000 times, and report the average processing time on different hardware platforms in Table \ref{Tab:inferenceTime}. 

Specifically, for the privacy preserver consisting of four convolutional and up-convolutional blocks, the average processing time %of the deployed privacy preserver 
on the desktop with an NVIDIA GeForce RTX 3080Ti GPU is less than 4 ms. The processing time on the laptops with an NVIDIA GeForce RTX 3060 GPU and with an NVIDIA GeForce RTX 1050Ti GPU is 10.9 ms and 53.5 ms, respectively. 
When reducing the number of the convolutional and up-convolutional blocks to three, the average processing time decreases.  On the desktop, it is reduced to 2.7 ms, on the laptop with an NVIDIA GeForce RTX 3060 GPU it is 9.6 ms, and the laptop with an NVIDIA GeForce RTX 1050Ti GPU it is 46.7 ms. These results indicate that the deployed privacy preserver introduces minimal processing latency.

\begin{table}[]

\begin{minipage}[c]{0.46\textwidth}

\centering
%\captionsetup{font={small,stretch=1.}, justification=raggedright}
\caption{The processing time (in ms) on different hardware platforms. \name does not introduce too much processing latency.}
\label{Tab:inferenceTime}
\resizebox{0.82\textwidth}{!}{%
\begin{tabular}{ccc}
\Xhline{2\arrayrulewidth}
Platforms            & 3 Blocks  & 4 Blocks \\ \hline
Desktop (RTX 3080Ti) &2.7        &  3.8          \\
Laptop (RTX 3060)    &9.6        &  10.9          \\
Laptop (RTX 1050Ti)  &46.7       &  53.5          \\  \Xhline{2\arrayrulewidth}
\end{tabular}%
}
\end{minipage}
\quad
%
% \hfill
\begin{minipage}[c]{0.46\textwidth}
\centering
%\captionsetup{font={small,stretch=1.}, justification=raggedright}
\caption{The memory usage (in MB) on different hardware platforms. \name consumes similar memories on different hardware platforms.}
\label{Tab:memoryConsumption}
\resizebox{0.825\textwidth}{!}{%
\begin{tabular}{ccc}
\Xhline{2\arrayrulewidth}
Platforms            & 3 Blocks & 4 Blocks \\ \hline
Desktop (RTX 3080Ti) & 2193     & 2267         \\
Laptop (RTX 3060)    & 1964     & 2043         \\
Laptop (RTX 1050Ti)  & 1902     & 2003         \\ \Xhline{2\arrayrulewidth}
\end{tabular}%
}
\end{minipage}
% \vspace{-0.1in}
\end{table}

\begin{table}[]
\centering

\caption{The utility and privacy performance of \name in different structures.
}
\resizebox{0.68\textwidth}{!}{%
\begin{tabular}{cccc}
\Xhline{2\arrayrulewidth}
Task                                   &  3 Blocks      & 4 Blocks      & w/o Defense\\ \hline
Utility: Gaze estimation               &  8.12$^\circ$  & 7.23$^\circ$  & 7.28$^\circ$  \\
Privacy: Identity recognition accuracy &  6.36\%        & 6.31\%        & 99.8\%\\
Privacy: Gender recognition accuracy   &  62.0\%        & 62.0\%        & 99.4 \%\\ \Xhline{2\arrayrulewidth}
\end{tabular}
}
\vspace{-0.15in}
\label{Tab:DifferentStructure}
\end{table}

\vspace{0.05in}
\noindent\textbf{Memory usage.} To measure memory usage, we follow the method described in \cite{MobileDeepPill} by reporting the memory allocated specifically to the privacy preserver. This is determined by subtracting the memory usage before loading the privacy preserver from the run-time memory usage. The results are shown in Table \ref{Tab:memoryConsumption}, indicating similar memory usage across different scenarios, approximately 2,000 MB for the privacy preserver.

\vspace{0.05in}
\noindent\textbf{Utility and privacy performance.} We report the utility and privacy performance of the privacy preserver with different structures in Table \ref{Tab:DifferentStructure}. When reducing the number of convolutional and up-convolutional blocks, the utility performance of \name is decreased to 8.12$^{\circ}$, which is approximately $1^{\circ}$ higher than w/o defense, while the privacy performance maintains stable. Therefore, although using three convolutional and up-convolutional blocks can slightly reduce the processing time, we opt to design the privacy preserver with four convolutional and up-convolutional blocks to ensure comparable utility performance to w/o defense.

\subsection{Discussions}
\label{subsec:discussion}
Below, we discuss the key findings of this paper and the impacts of \name. We also discuss its limitations %of \name 
and propose future research directions for enhancing user privacy in black-box mobile services.

\vspace{0.08in}
\noindent\textbf{Key findings.} This work presents three major findings. First, we demonstrate effective user privacy preservation by transforming different raw images into obfuscated images that have similar appearances with a pre-generated anchor image. %, known as the anchor image. 
Second, leveraging the anchor image allows us to control the appearance of obfuscated images, thereby achieving our utility goal. Specifically, since the anchor image, i.e., the average full-face image, produces consistent outputs for both $\mathcal{G}_w(\cdot)$ and $\mathcal{G}_b(\cdot)$, obfuscated images that closely resemble the anchor image also yield consistent results for gaze estimators $\mathcal{G}_w(\cdot)$ and $\mathcal{G}_b(\cdot)$. This alignment enables $\mathcal{P}(\cdot)$ trained with $\mathcal{G}_w(\cdot)$ to perform accurate gaze estimation on $\mathcal{G}_b(\cdot)$. Lastly, our well-designed $\mathcal{P}(\cdot)$ structure and training objective allow us to manipulate the behaviours of gaze estimators through imperceptible modification applied to the anchor image. This finding underscores vulnerabilities in deep learning-based gaze estimation systems.

\vspace{0.08in}
\noindent\textbf{Impacts of \name.} Compared to existing works~\cite{wu2021dapter, Liu_PAN_2020, Francesco_LPPE_2019}, \name addresses a more practical scenario where the deep learning-based model used by the service provider remains a black box to users. \name introduces a novel framework designed to preserve user privacy while maintaining good gaze estimation performance on such black-box models. While our current evaluation focuses on preserving identity and gender as private attributes, the framework's flexibility allows for the preservation of other private attributes, such as ages, emotions, and details of the user's surroundings. Moreover, \name can be extended to preserve user privacy in various applications, including head pose estimation \cite{headpose} and emotion recognition \cite{emotionRecognition}, by adapting the utility goals accordingly.

\vspace{0.08in}
\noindent\textbf{Limitations.} Our experiments have demonstrated that \name outperforms DP-based methods in achieving both privacy and utility goals. However, it is important to note that unlike DP-based methods, the current design of \name does not provide a theoretical privacy guarantee.

\vspace{0.08in}
\noindent\textbf{Future research directions.} A promising avenue for future research involves extending \name to other applications, such as hand pose estimation \cite{headpose}  %This can be achieved by adapting utility goals %and assessing the effectiveness of \name in preserving additional user private attributes beyond identity and gender. 
Another intriguing direction is to develop privacy-preserving solutions tailored for wearable-based gaze estimation systems that either utilize near-eye pupil images\cite{10.1145/2638728.2641695,kim2019nvgaze} or event streams~\cite{angelopoulos2021event,zhao2024ev,zhang2024swift,bonazzi2024retina} as tracking inputs. These systems pose unique challenges due to the sensitivity of the data captured and the wide adoption of eye tracking in head-mounted platforms such as augmented/virtual reality devices~\cite{clay2019eye,plopski2022eye}. Designing effective privacy-preserving solutions for such systems could significantly enhance user trust and adoption in these technologies.

%% file: 07conclusion.tex
\section{Conclusion}
\label{sec:conlusion}
In this work, we present \name, the first approach that can effectively preserve users' privacy information when calling black-box gaze tracking services without compromising the estimation performance. \name trains a user-side privacy preserver to convert privacy-sensitive full-face images into privacy-enhanced obfuscated versions. The obfuscated images do not contain any information about the users' private attributes yet can be directly used by the black-box gaze estimator to obtain accurate gaze directions. Our comprehensive experiments on four benchmark datasets show that \name can effectively protect users' private attributes, e.g., identity and gender, even when the attribute recognizers are trained on obfuscated images with accurate attribute labels. Meanwhile, the obfuscated images generated by \name can achieve comparable gaze tracking performance to conventional, unprotected full-face images. 